\documentclass[journal]{IEEEtran}
\ifCLASSINFOpdf
  \usepackage[pdftex]{graphicx}
\else
\fi
\usepackage{amssymb}
\usepackage{amsmath}
\newtheorem{thm}{Theorem}

\usepackage{algorithm}
\usepackage{algorithmic}

\usepackage{cases}
\usepackage{array}

\usepackage[tight,footnotesize]{subfigure}
\begin{document}
%
\title{Heterogeneous Multi-task Metric Learning across Multiple Domains}
%
%
%

\author{Yong~Luo,
        Yonggang~Wen,~\IEEEmembership{Senior Member,~IEEE,}
        and~Dacheng~Tao,~\IEEEmembership{Fellow,~IEEE}
\thanks{Y. Luo and Y. Wen are with the School of Computer Science and Engineering, Nanyang Technological University, Singapore 639798, e-mail: yluo180@gmail.com, ygwen@ntu.edu.sg.}
\thanks{D. Tao is with the Centre for Quantum Computation \& Intelligent Systems and the Faculty of Engineering and Information Technology, University of Technology, Sydney, 81 Broadway Street, Ultimo, NSW 2007, Australia, e-mail: dacheng.tao@uts.edu.au.}
\thanks{\copyright 2017 IEEE. Personal use of this material is permitted. Permission from IEEE must be obtained for all other uses, in any current or future media, including reprinting/republishing this material for advertising or promotional purposes, creating new collective works, for resale or redistribution to servers or lists, or reuse of any copyrighted component of this work in other works.}
}


%
%

\markboth{$>$ \normalsize{TNNLS-2015-P-5907 R}\footnotesize{evision}\normalsize{ 4} $<$}%
{Shell \MakeLowercase{\textit{et al.}}: Bare Demo of IEEEtran.cls for IEEE Journals}

%



\maketitle

\begin{abstract}
Distance metric learning (DML) plays a crucial role in diverse machine learning algorithms and applications. When the labeled information in target domain is limited, transfer metric learning (TML) helps to learn the metric by leveraging the sufficient information from other related domains. Multi-task metric learning (MTML), which can be regarded as a special case of TML, performs transfer across all related domains. Current TML tools usually assume that the same feature representation is exploited for different domains. However, in real-world applications, data may be drawn from heterogeneous domains. Heterogeneous transfer learning approaches can be adopted to remedy this drawback by deriving a metric from the learned transformation across different domains. But they are often limited in that only two domains can be handled. To appropriately handle multiple domains, we develop a novel heterogeneous multi-task metric learning (HMTML) framework. In HMTML, the metrics of all different domains are learned together. The transformations derived from the metrics are utilized to induce a common subspace, and the high-order covariance among the predictive structures of these domains is maximized in this subspace. There do exist a few heterogeneous transfer learning approaches that deal with multiple domains, but the high-order statistics (correlation information), which can only be exploited by simultaneously examining all domains, is ignored in these approaches. Compared with them, the proposed HMTML can effectively explore such high-order information, thus obtaining more reliable feature transformations and metrics. Effectiveness of our method is validated by the extensive and intensive experiments on text categorization, scene classification, and social image annotation.
\end{abstract}


%
\IEEEpeerreviewmaketitle

\section{Introduction}
\label{sec:Introduction}

%
%
%
%

%
%

\IEEEPARstart{T}{he} objective of distance metric learning (DML) is to find a measure to appropriately estimate the distance or similarity between data. DML is critical in various research fields, e.g., k-means clustering, k-nearest neighbor (kNN) classification, the sophisticated support vector machine (SVM) \cite{ZX-Xu-et-al-arXiv-2013, P-Bouboulis-et-al-TNNLS-2015} and learning to rank \cite{B-McFee-and-GR-Lanckriet-ICML-2010}. For instance, the kNN based models can be comparable or superior to other well-designed classifiers by learning proper distance metrics \cite{KQ-Weinberger-et-al-NIPS-2005, M-Guillaumin-et-al-ICCV-2009}. Besides, it is demonstrated in \cite{ZX-Xu-et-al-arXiv-2013} that learning the Mahalanobis metric for the RBF kernel in SVM without model selection consistently outperforms the traditional SVM-RBF, in which the hyperparameter is determined by cross validation.

Recently, some transfer metric learning (TML) \cite{ZJ-Zha-et-al-IJCAI-2009, Y-Zhang-and-DY-Yeung-TIST-2012} methods are proposed for DML in case that labeled data provided in target domain (the domain of interest) is insufficient, while the labeled information in certain related, but different source domains is abundant. In this scenario, the data distributions of the target and source domain may differ a lot, hence the traditional DML algorithms usually do not perform well. Indeed, TML \cite{ZJ-Zha-et-al-IJCAI-2009, Y-Zhang-and-DY-Yeung-TIST-2012} tries to reduce the distribution gap and help learn the target metric by utilizing the labeled data from the source domains. In particular, multi-task metric learning (MTML) \cite{Y-Zhang-and-DY-Yeung-TIST-2012} aims to simultaneously improve the metric learning of all different (source and target) domains since the labeled data for each of them is assumed to be scarce.

There is a main drawback in most of the current TML algorithms, i.e., samples of related domains are assumed to have the same feature dimension or share the same feature space. This assumption, however, is invalid in many applications. A typical example is the classification of multilingual documents. In this application, the feature representation of a document written in one language is different from that written in other languages since different vocabularies are utilized. Also, in some face recognition and object recognition applications, images may be collected under different environmental conditions, hence their representations have different dimensions. Besides, in natural image classification and multimedia retrieval, we often extract different kinds of features (such as global wavelet texture and local SIFT \cite{DG-Lowe-IJCV-2004}) to represent the samples, which can also have different modalities (such as text, audio and image). Each feature space or modality can be regarded as a domain.

In the literature, various heterogeneous transfer learning \cite{XX-Shi-et-al-ICDM-2010, C-Wang-and-S-Mahadevan-IJCAI-2011, JT-Zhou-et-al-AISTATS-2014} algorithms have been developed to deal with heterogeneous domains. In these approaches, a widely adopted strategy is to reduce the difference of heterogeneous domains \cite{JT-Zhou-et-al-AISTATS-2014} by transforming their representations into a common subspace. A metric can be derived from the transformation learned for each domain. Although has achieve success in some applications, these approaches suffer a major limitation that only two domains (one source and one target domain) can be handled. In practice, however, the number of domains is usually more than two. For instance, the news from the Reuters multilingual collection is written in five different languages. Besides, it is common to utilize various kinds of features (such as global, local, and biologically inspired) in visual analysis-based applications.

To remedy these drawbacks, we propose a novel heterogeneous multi-task metric learning (HMTML) framework, which is able to handle an arbitrary number of domains. In the proposed HMTML, the metrics of all domains are learned in a single optimization problem, where the empirical losses w.r.t. the metric for each domain are minimized. Meanwhile, we derive feature transformations from the metrics and use them to project the predictive structures of different domains into a common subspace. In this paper, all different domains are assumed to have the same application \cite{C-Wang-and-S-Mahadevan-IJCAI-2011, Y-Zhang-and-DY-Yeung-AAAI-2011}, such as article classification with the same categories. Consequently, the predictive structures, which are parameterized by the weight vector of classifiers, should be close to each other in the subspace. Then a connection between different metrics is built by minimizing the divergence between the transformed predictive structures. Compared to the learning of different metrics separately, more reliable metrics can be obtained since different domains help each other in our method. This is particularly important for those domains with little label information.

There do exist a few approaches \cite{C-Wang-and-S-Mahadevan-IJCAI-2011, Y-Zhang-and-DY-Yeung-AAAI-2011} that can learn transformations and derive metrics for more than two domains. These approaches, however, only explore the statistics (correlation information) between pairs of representations in either one-vs-one \cite{C-Wang-and-S-Mahadevan-IJCAI-2011}, or centralized \cite{Y-Zhang-and-DY-Yeung-AAAI-2011} way. While the high-order statistics are ignored, which can only be obtained by examining all domains simultaneously. Our method is more advantageous than these approaches in that we directly analyze the covariance tensor over the prediction weights of all domains. This encodes the high-order correlation information in the learned transformations and thus hopefully we can achieve better performance. Experiments are conducted extensively on three popular applications, i.e., text categorization, scene classification, and social image annotation. We compare the proposed HMTML with not only the Euclidean (EU) baseline and representative DML (single domain) algorithms \cite{KQ-Weinberger-et-al-NIPS-2005, JV-Davis-et-al-ICML-2007, R-Jin-et-al-NIPS-2009}, but also two representative heterogeneous transfer learning approaches \cite{C-Wang-and-S-Mahadevan-IJCAI-2011, Y-Zhang-and-DY-Yeung-AAAI-2011} that can deal with more than two domains. The experimental results demonstrate the superiority of our method.

This paper differs from our conference works \cite{Y-Luo-et-al-IJCAI-2016, Y-Luo-et-al-IJCAI-2017} significantly since: 1) in \cite{Y-Luo-et-al-IJCAI-2016}, we need large amounts of unlabeled corresponding data to build domain connection. In this paper, we utilize the predictive structures to bridge different domains and thus only limited labeled data are required. Therefore, the critical regularization term that enables knowledge transfer is quite different from that in \cite{Y-Luo-et-al-IJCAI-2016}; 2) in \cite{Y-Luo-et-al-IJCAI-2017}, we aim to learn features for different domains, instead of the distance metrics in this paper. The optimization problem is thus quite different since we directly optimize w.r.t. the distance metrics in this paper.

The rest of the paper is structured as follows. Some closely related works are summarized in Section \ref{sec:Related_Work}. In Section \ref{sec:HMTML}, we first give an overall description of the proposed HMTML, and then present its formulation together with some analysis. Experimental results are reported and analyzed in Section \ref{sec:Experiments}, and we conclude this paper in Section \ref{sec:Conclusion}.




\section{Related Work}
\label{sec:Related_Work}

\subsection{Distance metric learning}

The goal of distance metric learning (DML) \cite{XX-Xu-et-al-TNNLS-2015} is to learn an appropriate distance function over the input space, so that the relationships between data are appropriately reflected. Most conventional metric learning methods, which are often called ``Mahalanobis metric learning'', can be regarded as learning a linear transformation of the input data \cite{B-Kulis-FTML-2012, A-Bellet-and-AHM-Sebban-arXiv-2014}. The first work of Mahalanobis metric learning was done by Xing et al. \cite{EP-Xing-et-al-NIPS-2002}, where a constrained convex optimization problem with no regularization was proposed. Some other representative algorithms include the neighborhood component analysis (NCA) \cite{J-Goldberger-et-al-NIPS-2004}, large margin nearest neighbors (LMNN) \cite{KQ-Weinberger-et-al-NIPS-2005}, information theoretic metric learning (ITML) \cite{JV-Davis-et-al-ICML-2007}, and regularized distance metric learning (RDML) \cite{R-Jin-et-al-NIPS-2009}.

To capture the nonlinear structure in the data, one can extend the linear metric learning methods to learn nonlinear transformations by adopting the kernel trick or ``Kernel PCA (KPCA)'' trick \cite{R-Chatpatanasiri-et-al-NEUCOM-2010}. Alternatively, neural networks can be utilized to learn arbitrarily complex nonlinear mapping for metric learning \cite{S-Chopra-et-al-CVPR-2005}. Some other representative nonlinear metric learning approaches include the gradient-boosted LMNN (GB-LMNN) \cite{D-Kedem-et-al-NIPS-2012}, Hamming distance metric learning (HDML) \cite{M-Norouzi-et-al-NIPS-2012}, and support vector metric learning (SVML) \cite{ZX-Xu-et-al-arXiv-2013}.

Recently, transfer metric learning (TML) has attracted intensive attention to tackle the labeled data deficiency issue in the target domain \cite{B-Geng-et-al-TIP-2011, Y-Zhang-and-DY-Yeung-TIST-2012} or all given related domains \cite{S-Parameswaran-and-KQ-Weinberger-NIPS-2010, PP-Yang-et-al-MLJ-2013, Y-Zhang-and-DY-Yeung-TIST-2012}. The latter is often called multi-task metric learning (MTML), and is the focus of this paper. In \cite{S-Parameswaran-and-KQ-Weinberger-NIPS-2010}, they propose mt-LMNN, which is an extension of LMNN to the multi-task setting \cite{C-Li-et-al-TNNLS-2015}. This is realized by representing the metric of each task as a summation of a shared Mahalanobis metric and a task-specific metric. The level of information to be shared across tasks is controlled by the trade-off hyper-parameters of the metric regularizations. Yang et al. \cite{PP-Yang-et-al-MLJ-2013} generalize mt-LMNN by utilizing von Neumann divergence based regularization to preserve geometry between the learned metrics. An implicit assumption of these methods is that the data samples of different domains lie in the same feature space. Thus these approaches cannot handle heterogeneous features. To remedy this drawback, we propose heterogeneous MTML (HMTML) inspired by heterogeneous transfer learning.

\subsection{Heterogeneous transfer learning}

Developments in transfer learning \cite{L-Shao-et-al-TNNLS-2015} across heterogeneous feature spaces can be grouped to two categories: heterogeneous domain adaptation (HDA) \cite{C-Wang-and-S-Mahadevan-IJCAI-2011, JT-Zhou-et-al-AISTATS-2014} and heterogeneous multi-task learning (HMTL) \cite{Y-Zhang-and-DY-Yeung-AAAI-2011}. In HDA, there is usually only one target domain that has limited labeled data, and our aim is to utilize sufficient labeled data from related source domains to help the learning in the target domain. Whereas in HMTL, the labeled data in all domains are scarce, and thus we treat different domains equally and make them help each other.

Most HDA methods only have two domains, i.e., one source and one target domain. The main idea in these methods is to either map the heterogeneous data into a common feature space by learning a feature mapping for each domain \cite{XX-Shi-et-al-ICDM-2010, LX-Duan-et-al-ICML-2012}, or map the data from the source domain to the target domain by learning an asymmetric transformation \cite{B-Kulis-et-al-CVPR-2011, JT-Zhou-et-al-AISTATS-2014}. The former is equivalent to Mahalanobis metric learning since each learned mapping could be used to derive a metric directly. Wang and Mahadevan \cite{C-Wang-and-S-Mahadevan-IJCAI-2011} presented a HDA method based on manifold alignment (DAMA). This method manages several domains, and the feature mappings are learned by utilizing the labels shared by all domains to align each pair of manifolds. Compared with HDA, there are much fewer works on HMTL. One representative approach is the multi-task discriminant analysis (MTDA) \cite{Y-Zhang-and-DY-Yeung-AAAI-2011}, which extends linear discriminant analysis (LDA) to learn multiple tasks simultaneously. In MTDA, a common intermediate structure is assumed to be shared by the learned latent representations of different domains. MTDA can also deal with more than two domains, but is limited in that only the pairwise correlations (between each latent representation and the shared representation) are exploited. Therefore, the high-order correlations between all domains are ignored in both DAMA and MTDA. We develop the following tensor based heterogeneous multi-task metric learning framework to rectify this shortcoming.

\begin{figure*}[!t]
\centering
\includegraphics[width=1.4\columnwidth]{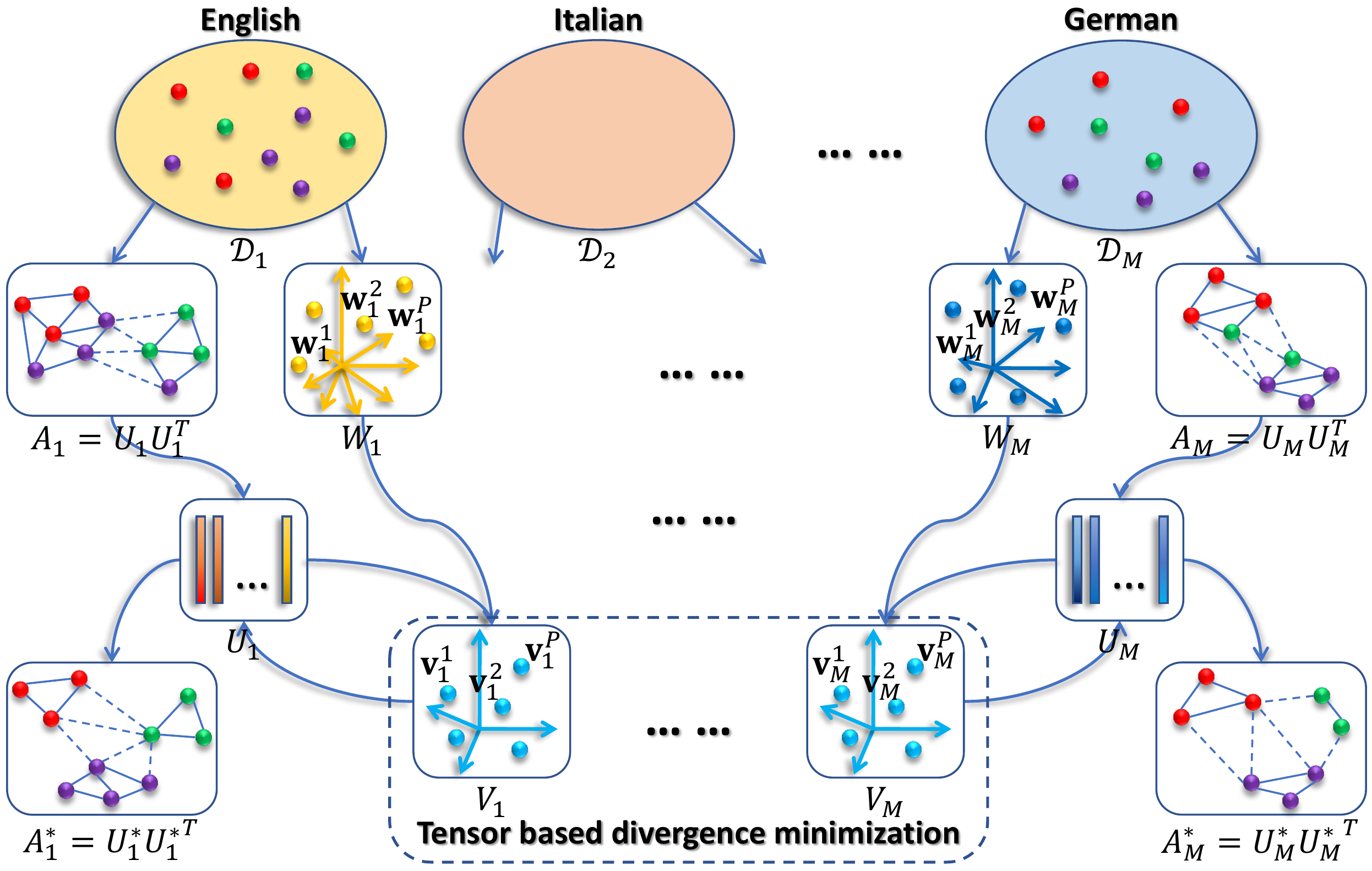}
\caption{System diagram of the proposed heterogeneous multi-task metric learning. Given limited labeled samples for each of the $M$ domains, learning the metrics $A_1, A_2, \ldots, A_M$ separately may be unreliable. To allow all different domains help each other, we firstly learn the prediction weights $W_m = [\mathbf{w}_m^1, \mathbf{w}_m^2, \ldots, \mathbf{w}_m^P], m = 1, 2, \ldots, M$. Considering the metric $A_m$ can be decomposed as $A_m = U_m U_m^T$, we transform all $\{W_m\}$ to a common subspace using $\{ U_m \}$ and obtain $\{ V_m = U_m^T W_m \}$, where $V_m = [\mathbf{v}_m^1, \mathbf{v}_m^2, \ldots, \mathbf{v}_m^P]$ with each $\mathbf{v}_m^p = U_m^T \mathbf{w}_m^p$. By simultaneously minimizing the empirical losses w.r.t. each $A_m$ and the high-order divergence between all $\{ V_m \}$, we obtain reliable $U_m^\ast$ and also the metric $A_m^\ast = U_m^\ast (U_m^\ast)^T$.}
\label{fig:System_Diagram}
\end{figure*}

\section{Heterogeneous multi-task metric learning}
\label{sec:HMTML}

In DAMA \cite{C-Wang-and-S-Mahadevan-IJCAI-2011} and MTDA \cite{Y-Zhang-and-DY-Yeung-AAAI-2011}, linear transformation is learned for each domain and only pairwise correlation information is explored. In contrast to them, tensor based heterogeneous MTML (HMTML) is developed in this paper to exploit the high-order tensor correlations for metric learning. Fig. \ref{fig:System_Diagram} is an overview of the proposed method. Here, we take multilingual text classification as an example to illustrate the main idea. The number of labeled samples for each of the $M$ heterogenous domains (e.g., ``English'', ``Italian'', and ``German'') are assumed to be small. For the $m$'th domain, empirical losses w.r.t. the metric $A_m$ are minimized on the labeled set $\mathcal{D}_m$. The obtained metrics may be unreliable due to the labeled data deficiency issue in each domain. Hence we let the different domains help each other to learn improved metrics by sharing information across them. This is performed by first constructing multiple prediction (such as binary classification) problems in the $m$'th domain, and a set of prediction weight vectors $\{\mathbf{w}_m^p\}_{p=1}^P$ are obtained by training the problems on $\mathcal{D}_m$. Here, $\mathbf{w}_m^p$ is the weight vector of the $p$'th base classifier, and we assume each of the $P$ base classifiers is linear, i.e., $f^p(\mathbf{x}_m) = (\mathbf{w}_m^p)^T \mathbf{x}_m$. We follow \cite{JT-Zhou-et-al-AISTATS-2014} to generate the $P$ prediction problems, where the well-known Error Correcting Output Codes (ECOC) scheme is adopted. Then we decompose the metric $A_m$ as $A_m=U_m U_m^T$, and the obtained $U_m$ is used to transform the weight vectors into a common subspace, i.e., $\{\mathbf{v}_m^p = U_m^T \mathbf{w}_m^p\}_{p=1}^P$. As illustrated in Section \ref{sec:Introduction}, the application is the same for different domains. Therefore, after being transformed into the common subspace, the weight vectors should be close to each other, i.e., all $\{ \mathbf{v}_1^p, \mathbf{v}_2^p, \ldots, \mathbf{v}_m^p \}$ should be similar. Finally, by maximizing the tensor based high-order covariance between all $\{ \mathbf{v}_m^p \}$, we obtain improved $U_m^\ast$, where additional information from other domains are involved. Then more reliable metric can be obtained since $A_m^\ast = U_m^\ast (U_m^\ast)^T$.

In short, ECOC generates a binary ``codebook'' to encode each class as a binary string, and trains multiple binary classifiers according to the ``codebook''. The test sample is classified using all the trained classifiers and the results are concatenated to obtain a binary string, which is decoded for final prediction \cite{TG-Dietterich-and-G-Bakiri-JAIR-1995}. In this paper, the classification weights of all the trained binary classifiers are used to build domain connections. It should be noted that the learned weights may not be reliable if given limited labeled samples. Fortunately, we can construct sufficient base classifiers using the ECOC scheme, and thus robust transformations can be obtained even some learned base classifiers are inaccurate or incorrect \cite{JT-Zhou-et-al-AISTATS-2014}. The reasons why the decomposition $U_m$ of the metric $A_m$ can be used to transform the parameters of classifiers are mainly based on two points: 1) Mahanobias metric learning can be formulated as learning a linear transformation \cite{B-Kulis-FTML-2012}. In the literature of distance metric learning (DML), most methods focus on learning the Mahalanobis distance, which is often denoted as
\begin{equation}
\notag
d_A (\mathbf{x}_i, \mathbf{x}_j) = (\mathbf{x}_i - \mathbf{x}_j)^T A (\mathbf{x}_i - \mathbf{x}_j),
\end{equation}
where $A$ is the metric, which is a positive semi-definite matrix and can be factorized as $A = U U^T$. By applying some simple algebraic manipulations, we have $d_A (\mathbf{x}_i, \mathbf{x}_j ) = \| U \mathbf{x}_i - U \mathbf{x}_j \|_2^2$; 2) according to the multi-task learning methods presented in \cite{RK-Ando-and-T-Zhang-JMLR-2005, A-Quattoni-et-al-CVPR-2007}, the transformation learned in the parameter space can also be interpreted as defining mapping in the feature space. Therefore, the linear transformation $U_m$ can be used to transform the parameters of classifiers. The technical details of the proposed method are given below, and we start by briefing the frequently used notations and concepts of multilinear algebra in this paper.

\subsection{Notations}
If $\mathcal{A}$ is an $M$-th order tensor of size $I_1 \times I_2 \times \ldots \times I_M$, and $U$ is a $J_m \times I_m$ matrix, then the $m$-mode product of $\mathcal{A}$ and $U$ is signified as $\mathcal{B} = \mathcal{A} \times_m U$, which is also an $M$-th order tensor of size $I_1 \times \ldots \times I_{m-1} \times J_m \times I_{m+1} \ldots \times I_M$ with the entry
\begin{equation}
\label{eq:TTM}
\begin{split}
& \mathcal{B}(i_1, \ldots, i_{m-1}, j_m, i_{m+1}, \ldots, i_M) \\
= & \sum_{i_m=1}^{I_m} \mathcal{A}(i_1, i_2, \ldots, i_M) U(j_m, i_m).
\end{split}
\end{equation}
The product of $\mathcal{A}$ and a set of matrices $\{ U_m \in \mathbb{R}^{J_m \times I_m} \}_{m=1}^M$ is given by
\begin{equation}
\label{eq:TTMS}
\mathcal{B} = \mathcal{A} \times_1 U_1 \times_2 U_2 \ldots \times_M U_M.
\end{equation}
The mode-$m$ matricization of $\mathcal{A}$ is a matrix $A_{(m)}$ of size $I_m \times (I_1 \ldots I_{m-1} I_{m+1} \ldots I_M)$. We can regard the $m$-mode multiplication $\mathcal{B} = \mathcal{A} \times_m U$ as matrix multiplication in the form of $B_{(m)} = U A_{(m)}$.
Let $\mathbf{u}$ be an $I_m$-vector, the contracted $m$-mode product of $\mathcal{A}$ and $\mathbf{u}$ is denoted as $\mathcal{B} = \mathcal{A} \bar{\times}_m \mathbf{u}$, which is an $M-1$-th tensor of size $I_1 \times \ldots \times I_{m-1} \times I_{m+1} \ldots \times I_M$. The elements are calculated by
\begin{equation}
\begin{split}
& \mathcal{B}(i_1, \ldots, i_{m-1}, i_{m+1}, \ldots, i_M)
= \sum_{i_m=1}^{I_m} \mathcal{A}(i_1, i_2, \ldots, i_M) \mathbf{u}(i_m).
\end{split}
\end{equation}
Finally, the Frobenius norm of the tensor $\mathcal{A}$ is calculated as
\begin{equation}
\| \mathcal{A} \|_F^2 = \langle \mathcal{A}, \mathcal{A} \rangle = \sum_{i_1=1}^{I_1} \sum_{i_2=1}^{I_2} \ldots \sum_{i_M=1}^{I_M} \mathcal{A}(i_1, i_2, \ldots, i_M)^2.
\end{equation}

\subsection{Problem formulation}

Suppose there are $M$ heterogeneous domains, and the labeled training set for the $m$'th domain is $\mathcal{D}_m = \{(\mathbf{x}_{mn}, y_{mn})\}_{n=1}^{N_m}$, where $\mathbf{x}_{mn} \in \mathbb{R}^{d_m}$ and its corresponding class label $y_{mn} \in \{1, 2, \ldots, C\}$. The label set is the same for the different heterogeneous domains \cite{C-Wang-and-S-Mahadevan-IJCAI-2011, Y-Zhang-and-DY-Yeung-AAAI-2011}. Then we have the following general formulation for the proposed HMTML
\begin{equation}
\label{eq:HMTML_General}
\begin{split}
\mathop{\min}_{\{A_m\}_{m=1}^M} & \ F(\{A_m\}) = \sum_{m=1}^M \Psi(A_m) + \gamma R(A_1, A_2, \ldots, A_M), \\
\mathrm{s.t.} & \ A_m \succeq 0, m = 1, 2, \ldots, M,
\end{split}
\end{equation}
where $\Psi(A_m) = \frac{2}{N_m (N_m-1)} \sum_{i<j} L(A_m; \mathbf{x}_{mi}, \mathbf{x}_{mj}, y_{mij})$  is the empirical loss w.r.t. $A_m$ in the $m$'th domain, $y_{mij} = \pm 1$ indicates $\mathbf{x}_{mi}$ and $\mathbf{x}_{mj}$ are similar/dissimilar to each other, and $R(A_1, A_2, \ldots, A_M)$ is a carefully chosen regularizer that enables knowledge transfer across different domains. Here, $y_{mij}$ is obtained according to whether $\mathbf{x}_{mi}$ and $\mathbf{x}_{mj}$ belong to the same class or not. Following \cite{R-Jin-et-al-NIPS-2009}, we choose $L(A_m; \mathbf{x}_{mi}, \mathbf{x}_{mj}, y_{mij}) = g(y_{mij} [1 - \|\mathbf{x}_{mi} - \mathbf{x}_{mj}\|_{A_m}^2])$ and adopt the generalized log loss (GL-loss) \cite{RS-Cabral-et-al-NIPS-2011} for $g$, i.e., $g(z) = \frac{1}{\rho} \mathrm{log}(1+\mathrm{exp}(-\rho z))$. Here, $\|\mathbf{x}_{mi} - \mathbf{x}_{mj}\|_{A_m}^2 = (\mathbf{x}_{mi} - \mathbf{x}_{mj})^T A_m (\mathbf{x}_{mi} - \mathbf{x}_{mj})$. The GL-loss is a smooth version of the hinge loss, and the smoothness is controlled by the hyper-parameter $\rho$, which is set as $3$ in this paper. For notational simplicity, we denote $\mathbf{x}_{mi}$, $\mathbf{x}_{mj}$ and $y_{mij}$ as $\mathbf{x}_{mk}^1$, $\mathbf{x}_{mk}^2$ and $y_{mk}$ respectively, where $k = 1, 2, \ldots, N_m' = \frac{N_m (N_m-1)}{2}$. We also set $\delta_{mk} = \mathbf{x}_{mk}^1 - \mathbf{x}_{mk}^2$ so that $\|\mathbf{x}_{mk}^1 - \mathbf{x}_{mk}^2\|_{A_m}^2 = \delta_{mk}^T A_m \delta_{mk}$, and the loss term becomes $\Psi(A_m) = \frac{1}{N_m'} \sum_{k=1}^{N_m'} g\left( y_{mk}(1 - \delta_{mk}^T A_m \delta_{mk}) \right)$.

To transfer information across different domains, $P$ binary classification problems are constructed and a set of classifiers $\{ \mathbf{w}_m^p \}_{p=1}^P$ are learned for each of the $M$ domains by utilizing the labeled training data. This process can be carried out in an offline manner and thus does not has influence on the computational cost of subsequent metric learning. Considering that $A_m = U_m U_m^T$ due to the positive semi-definite property of the metric, we propose to use $U_m$ to transform $\mathbf{w}_m^p$ and this leads to $\mathbf{v}_m^p = U_m^T \mathbf{w}_m^p$. Then the divergence of all $\{ \mathbf{v}_1^p, \mathbf{v}_2^p, \ldots, \mathbf{v}_M^p \}$ are minimized. In the following, we first derive the formulation for $M = 2$ (two domains), and then generalize it for $M > 2$.

When $M = 2$, we have the following formulation:
\begin{equation}
\label{eq:HMTML_REG_2D}
\begin{split}
\mathop{\min}_{U_1, U_2} \frac{1}{P} & \ \sum_{p=1}^P \|\mathbf{v}_1^p - \mathbf{v}_2^p\|_2^2 + \sum_{m=1}^{2} \gamma_m \|U_m\|_1, \\
\mathrm{s.t.} & \ U_1, U_2 \succeq 0,
\end{split}
\end{equation}
where $U_m \in \mathbb{R}^{d_m \times r}$ is the transformation of the $m$'th domain, and $r$ is the number of common factors shared by different domains. The trade-off hyper-parameters $\{ \gamma_m > 0\}$, the $l_1$-norm $\| U_m \|_1 = \sum_{i=1}^d \sum_{j=1}^r |u_{mij}|$, and $\succeq$ indicates that all entries of $U_m$ are non-negative. The transformation $U_m$ is restricted to be sparse since the features in different domains usually have sparse correspondences. The non-negative constraints can narrow the hypothesis space and improve the interpretability of the results.

For $M > 2$, we generalize (\ref{eq:HMTML_REG_2D}) as
\begin{equation}
\label{eq:HMTML_REG_2D_Reformulation}
\begin{split}
\mathop{\min}_{U_1, U_2} \frac{1}{P} & \ \sum_{p=1}^P \|\mathbf{w}_1^p - G \mathbf{w}_2^p\|_2^2 + \sum_{m=1}^{2} \gamma_m \| U_m \|_1, \\
\mathrm{s.t.} & \ U_1, U_2 \succeq 0,
\end{split}
\end{equation}
where $G = U_1 E_r U_2^T $, and $E_r$ is an identity matrix of size $r$. By using the tensor notation, we have $G = E_r \times_1 U_1 \times_2 U_2$, so the formulation (\ref{eq:HMTML_REG_2D_Reformulation}) for $M > 2$ is given by
\begin{equation}
\label{eq:HMTML_REG}
\begin{split}
\mathop{\min}_{\{U_m\}_{m=1}^M} & \ \frac{1}{P} \sum_{p=1}^P \| \mathbf{w}_1^p - \mathcal{G} \bar{\times}_2 (\mathbf{w}_2^p)^T \ldots \bar{\times}_M (\mathbf{w}_M^p)^T \|_2^2 \\
& + \sum_{m=1}^M \gamma_m \|U_m\|_1, \\
\mathrm{s.t.} & \ U_m \succeq 0, m = 1, \ldots, M,
\end{split}
\end{equation}
where $\mathcal{G} = \mathcal{E}_r \times_1 U_1 \times_2 U_2 \ldots \times_M U_M$ is a transformation tensor, $\mathcal{E}_r \in \mathbb{R}^{r \times r \times \ldots \times r}$ is an identity tensor (the diagonal elements are $1$, and all other entries are $0$). A specific optimization problem for HMTML can be obtained by choosing the regularizer in (\ref{eq:HMTML_General}) as the objective of (\ref{eq:HMTML_REG}), i.e.,
\begin{equation}
\label{eq:HMTML_Specific}
\begin{split}
\mathop{\min}_{\{U_m\}_{m=1}^M} \ & F(\{U_m\}) \\
& = \sum_{m=1}^M \frac{1}{N_m'} \sum_{k=1}^{N_m'} g\left( y_{mk}(1 - \delta_{mk}^T U_m U_m^T \delta_{mk}) \right) \\
& + \frac{\gamma}{P} \sum_{p=1}^P \| \mathbf{w}_1^p - \mathcal{G} \bar{\times}_2 (\mathbf{w}_2^p)^T \ldots \bar{\times}_M (\mathbf{w}_M^p)^T \|_2^2 \\
& + \sum_{m=1}^M \gamma_m \|U_m\|_1, \\
& \mathrm{s.t.} \ U_m \succeq 0, m = 1, 2, \ldots, M.
\end{split}
\end{equation}
We reformulate (\ref{eq:HMTML_Specific}) using the following theorem due to its inconvenience in optimization.
\begin{thm}
\label{thm:Transform_Equivalence}
If $\| \mathbf{w}_m^p \|_2^2 = 1, p = 1, \ldots, P; m = 1, \ldots, M$, then we have:
\begin{equation}
\label{eq:Transform_Equivalence}
\begin{split}
& \| \mathbf{w}_1^p - \mathcal{G} \bar{\times}_2 (\mathbf{w}_2^p)^T \ldots \bar{\times}_M (\mathbf{w}_M^p)^T \|_2^2 \\
& = \| \mathbf{w}_1^p \circ \mathbf{w}_2^p \ldots \circ \mathbf{w}_M^p - \mathcal{G} \|_F^2,
\end{split}
\end{equation}
where $\circ$ is the outer product.
\end{thm}
We leave the proof in the supplementary material. 

By substituting (\ref{eq:Transform_Equivalence}) into (\ref{eq:HMTML_Specific}) and replacing $\mathcal{G}$ with $\mathcal{E}_r \times_1 U_1 \times_2 U_2 \ldots \times_M U_M$, we obtain the following reformulation of (\ref{eq:HMTML_Specific}):
\begin{equation}
\label{eq:HMTML_Reformulation}
\begin{split}
\mathop{\min}_{\{U_m\}_{m=1}^M} \ & F(\{U_m\}) \\
& = \sum_{m=1}^M \frac{1}{N_m'} \sum_{k=1}^{N_m'} g\left( y_{mk}(1 - \delta_{mk}^T U_m U_m^T \delta_{mk}) \right) \\
& + \frac{\gamma}{P} \sum_{p=1}^P \| \mathcal{W}^p - \mathcal{E}_r \times_1 U_1 \times_2 U_2 \ldots \times_M U_M \|_F^2 \\
& + \sum_{m=1}^M \gamma_m \|U_m\|_1, \\
& \mathrm{s.t.} \ U_m \succeq 0, m = 1, 2, \ldots, M,
\end{split}
\end{equation}
where $\mathcal{W}^p = \mathbf{w}_1^p \circ \mathbf{w}_2^p \ldots \circ \mathbf{w}_M^p$ is the covariance tensor of the prediction weights of all different domains. It is intuitive that a latent subspace shared by all domains can be found by minimizing the second term in (\ref{eq:HMTML_Reformulation}). In this subspace, the representations of different domains are close to each other and knowledge is transferred. Hence different domains can help each other to learn improved transformation $U_m$, and also the distance metric $A_m$.

\subsection{Optimization algorithm}
Problem (\ref{eq:HMTML_Reformulation}) can be solved using an alternating optimization strategy. That is, only one variable $U_m$ is updated at a time and all the other $U_{m'}$, $m' \neq m$ are fixed. This updating procedure is conducted iteratively for each variable. According to \cite{L-De-Lathauwer-et-al-JMAA-HOPM-2000}, we have
\begin{equation}
\notag
\begin{split}
& \mathcal{G} = \mathcal{E}_r \times_1 U_1 \times_2 U_2 \ldots \times_M U_M
 = \mathcal{B} \times_m U_m,
\end{split}
\end{equation}
where $\mathcal{B} = \mathcal{E}_r \times_1 U_1 \ldots \times_{m-1} U_{m-1} \times_{m+1} U_{m+1} \ldots \times_M U_M$. By applying the metricizing property of the tensor-matrix product, we have $G_{(m)} = U_m B_{(m)}$. Besides, it is easy to verify that $\| \mathcal{W}^p - \mathcal{G} \|_F^2 = \| W_{(m)}^p - G_{(m)} \|_F^2$. Therefore, the sub-problem of (\ref{eq:HMTML_Reformulation}) w.r.t. $U_m$ becomes:
\begin{equation}
\label{eq:HMTML_wrt_Um}
\begin{split}
\mathop{\min}_{U_m} & \ F(U_m) = \Phi(U_m) + \Omega(U_m), \\
\mathrm{s.t.} & \ U_m \succeq 0,
\end{split}
\end{equation}
where $\Phi(U_m) = \frac{1}{N_m'} \sum_{k=1}^{N_m'} g\left( y_{mk}(1 - \delta_{mk}^T U_m U_m^T \delta_{mk}) \right) + \frac{\gamma}{P} \sum_{p=1}^P \| W_{(m)}^p - U_m B_{(m)} \|_F^2$, and $\Omega(U_m) = \gamma_m \| U_m \|_1$. We propose to solve the problem (\ref{eq:HMTML_wrt_Um}) efficiently by utilizing the projected gradient method (PGM) presented in \cite{CJ-Lin-NCn-2007}. However, the term in $\Omega(U_m)$ is non-differentiable, we thus first smooth it according to \cite{Y-Nesterov-MP-2005}. For notational clarity, we omit the subscript $m$ in the following derivation. According to \cite{Y-Nesterov-MP-2005}, the smoothed version of the $l_1$-norm $l(u_{ij}) = |u_{ij}|$ can be given by
\begin{equation}
\label{eq:L1_Norm_Smooth}
l^\sigma(u_{ij}) = \mathop{\max}_{Q \in \mathcal{Q}} \langle u_{ij}, q_{ij} \rangle - \frac{\sigma}{2} q_{ij}^2,
\end{equation}
where $\mathcal{Q} = \{ Q: -1 \leq q_{ij} \leq 1, Q \in \mathbb{R}^{d \times r} \}$ and $\sigma$ is the smooth hyper-parameter, where we set it as $0.5$ empirically in our implementation according to the comprehensive study of the smoothed $l_1$-norm in \cite{TY-Zhou-et-al-ICDM-2010}. By setting the objective function of (\ref{eq:L1_Norm_Smooth}) to zero and then projecting $q_{ij}$ on $\mathcal{Q}$, we obtain the following solution, 
\begin{equation}
\label{eq:L1_Smooth_Solution}
q_{ij} = \mathrm{median} \left\{ \frac{u_{ij}}{\sigma}, -1, 1 \right\}.
\end{equation}
By substituting the solution (\ref{eq:L1_Smooth_Solution}) back into (\ref{eq:L1_Norm_Smooth}), we have the piece-wise approximation of $l$, i.e.,
\begin{equation}
\label{eq:L1_Norm_Approximation}
\begin{split}
l^\sigma= \left\{
\begin{array}{cc}
-u_{ij} - \frac{\sigma}{2}, & u_{ij} < -\sigma; \\
u_{ij} - \frac{\sigma}{2}, & u_{ij} > \sigma; \\
\frac{u_{ij}^2}{2 \sigma}, & \mathrm{otherwise}.
\end{array}
\right.
\end{split}
\end{equation}
To utilize the PGM for optimization, we have to compute the gradient of the smoothed $l_1$-norm to determine the descent direction. We summarize the results in the following theorem.
\begin{thm}
\label{thm:L1_Smooth_Gradient}
The gradient of smoothed $l(U) = \| U \|_1 = \sum_{i=1}^d \sum_{j=1}^r l(u_{ij})$ is
\begin{equation}
\label{eq:L1_Smooth_Gradient}
\frac{\partial l^\sigma(U)}{\partial U} = Q,
\end{equation}
where $Q$ is the matrix defined in (\ref{eq:L1_Norm_Smooth}).
\end{thm}
In addition, we summarize the gradient of $\Phi(U)$ as follows,
\begin{thm}
\label{thm:Phi_Gradient}
The gradient of $\Phi(U)$ w.r.t. $U$ is
\begin{equation}
\label{eq:Phi_Gradient}
\frac{\partial \Phi(U)}{\partial U} = \frac{1}{N'} \sum_k \frac{2 y_k (\delta_k \delta_k^T) U}{1 + \exp(\rho z_k)} + \frac{2 \gamma}{P} \sum_p (U B B^T - W^p B^T),
\end{equation}
where $z_k = y_k (1 - \delta_k^T U U^T \delta_k)$.
\end{thm}
We leave the proofs of both theorems in the supplementary material. Therefore, the gradient of the smoothed $F(U_m)$ is
\begin{equation}
\label{eq:F_Smooth_Gradient}
\begin{split}
& \frac{\partial F^\sigma(U_m)}{\partial U_m} = \frac{1}{N_m'} \sum_k \frac{2 y_{mk} (\delta_{mk} \delta_{mk}^T) U_m}{1 + \exp(\rho z_{mk})} \\
& + \frac{2 \gamma}{P} \sum_p \left( U_m B_{(m)} B_{(m)}^T - W_{(m)}^p B_{(m)}^T \right) + \gamma_m Q_m.
\end{split}
\end{equation}
Based on the obtained gradient, we apply the improved PGM presented in \cite{CJ-Lin-NCn-2007} to minimize the smoothed primal $F^\sigma(U_m)$, i.e.,
\begin{equation}
\label{eq:PGM_Update_Rule}
U_m^{t+1} = \pi[U_m^t - \mu_t \nabla F^\sigma(U_m^t)],
\end{equation}
where the operator $\pi[x]$ projects all the negative entries of $x$ to zero, and $\mu_t$ is the step size that must satisfy the following condition:
\begin{equation}
\label{eq:PGM_Step_Size}
F^\sigma(U_m^{t+1}) - F^\sigma(U_m^t) \leq \kappa \nabla F^\sigma(U_m^t)^T (U_m^{t+1} - U_m^t),
\end{equation}
where the hyper-parameter $\kappa$ is chosen to be $0.01$ following \cite{CJ-Lin-NCn-2007}. The step size can be determined using the Algorithm 1 cited from \cite{CJ-Lin-NCn-2007} (Algorithm 4 therein), and the convergence of the algorithm is guaranteed according to \cite{CJ-Lin-NCn-2007}. The stopping criterion we utilized here is $|F^\sigma(U_m^{t+1}) - F^\sigma(U_m^t)| / (|F^\sigma(U_m^{t+1}) - F^\sigma(U_m^0)| < \epsilon)$, where the initialization $U_m^0$ is the set as the results of the previous iterations in the alternating of all $\{U_m\}_{m=1}^M$.

Finally, the solutions of (\ref{eq:HMTML_Reformulation}) are obtained by alternatively updating each $U_m$ using Algorithm \ref{alg:Optimization_Um} until the stop criterion $|OBJ_{k+1} - OBJ_k| / |OBJ_k| < \epsilon$ is reached, where $OBJ_k$ is the objective value of (\ref{eq:HMTML_Reformulation}) in the $k$'th iteration step. Because the objective value of (\ref{eq:HMTML_wrt_Um}) decreases at each iteration of the alternating procedure, i.e., $F(U_m^{k+1}, \{U_{m'}^k\}_{m' \neq m}) \leq F(\{ U_m^k \})$. This indicates that $F(\{U_m^{k+1}\}) \leq F(\{ U_m^k \})$. Therefore, the convergence of the proposed HMTML algorithm is guaranteed. Once the solutions $\{U_m^\ast\}_{m=1}^M$ have been obtained, we can conduct subsequent learning, such as multi-class classification in each domain using the learned metric $A_m^\ast = U_m^\ast {U_m^\ast}^T$. More implementation details can be found in the supplementary material.

\begin{algorithm}[!t]
\caption{The improved projected gradient method for solving $U_m$.}
\label{alg:Optimization_Um}
\begin{algorithmic}[1]
\renewcommand{\algorithmicrequire}{\textbf{Input:}}
\REQUIRE Sample pair differences and corresponding labels $\{\delta_{mk}, y_{mk}\}_{k=1}^{N_m'}$; $B^{(m)}$ and $W_{(m)}^p, p = 1, \ldots, P$.
\renewcommand{\algorithmicrequire}{\textbf{Hyper-parameters:}}
\REQUIRE $\gamma$ and $\gamma_m$.
\renewcommand{\algorithmicensure}{\textbf{Output:}}
\ENSURE $U_m$.
\STATE{Set $\mu_0 = 1$, and $\beta = 0.1$. Initialize $U_m^0$ as the results of the previous iterations in the alternating optimization of all $\{U_m\}_{m=1}^M$.}
\renewcommand{\algorithmicrepeat}{\textbf{For $t = 1,2,\ldots$}}
\renewcommand{\algorithmicuntil}{\textbf{Until convergence}}
\REPEAT
\STATE{(a) Assign $\mu_t \leftarrow \mu_{t-1}$;}
\STATE{(b) \textbf{If} $\mu_t$ satisfies (\ref{eq:PGM_Step_Size}), repeatedly increase it by $\mu_t \leftarrow \mu_t/\beta$, until either $\mu_t$ does not satisfy (\ref{eq:PGM_Step_Size}) or $U_m(\mu_t/\beta) = U_m(\mu_t)$. \\
\textbf{Else} repeatedly decrease $\mu_t \leftarrow \mu_t*\beta$ until $\mu_t$ satisfy (\ref{eq:PGM_Step_Size});}
\STATE{(c) Update $U_m^{t+1} = \pi[U_m^t - \mu_t \nabla F^\sigma(U_m^t)]$.}
\UNTIL
\end{algorithmic}
\end{algorithm}

\subsection{Complexity analysis}

To analyze the time complexity of the proposed HMTML algorithm, we first present the computational cost of optimizing each $U_m$ as described in Algorithm \ref{alg:Optimization_Um}. In each iteration of Algorithm \ref{alg:Optimization_Um}, we must first determine the descent direction according to the gradient calculated using (\ref{eq:F_Smooth_Gradient}). Then an appropriate step size is obtained by exhaustedly checking whether the condition (\ref{eq:PGM_Step_Size}) is satisfied. In each check, we need to calculate the updated objective value of $F^\sigma(U_m^{t+1})$. The main cost of objective value calculation is spent on $\sum_{p=1}^P \| W_{(m)}^p - U_m B_{(m)} \|_F^2$, which can be rewritten as $\mathrm{tr}( ( U_m^T U_m B_{(m)} B_{(m)}^T ) - 2 \sum_{p=1}^P (W_{(m)}^p B_{(m)}^T ) U_m^T )$, where the constant part $\mathrm{tr}( \sum_{p=1}^P (W_{(m)}^p )^T W_{(m)}^p )$ is omitted. To accelerate computation, we pre-calculate $B_{(m)} B_{(m)}^T$ and $( \sum_p W_{(m)}^p ) B_{(m)}^T$, which are independent on $U_m$ and the time costs are $O( r^2 \prod_{m' \neq m} d_{m'} )$ and $O( r \prod_{m=1}^M d_m )$ respectively. After the pre-calculation, the time complexities of calculating $ (U_m^T U_m)(B_{(m)} B_{(m)}^T )$ and $\sum_{p=1}^P (W_{(m)}^p B_{(m)}^T) U_m^T$ become $O( r^2 d_m + r^{2.807} )$ and $O( r^2 d_m )$ respectively, where the complexity of $O( r^{2.807} )$ comes from the multiplication of two square matrices of size $r$ using the Strassen algorithm \cite{W-Miller-SJC-1975}. It is easy to derive that the computational cost of the remained parts in the objective function is $O( r d_m N_m' )$. Hence, the time cost of calculating the objective value is $O( r d_m N_m' + r^2 d_m + r^{2.807} )$ after the pre-calculation.

Similarly, the main cost of gradient calculation is spent on $\sum_p ( U_m B_{(m)} B_{(m)}^T - W_{(m)}^p B_{(m)}^T )$, and after pre-calculating $B_{(m)} B_{(m)}^T$ and $( \sum_p W_{(m)}^p ) B_{(m)}^T$, we can derive that the time cost of calculating the gradient is $O( r d_m N_m' + r^2 d_m )$. Therefore, the computational cost of optimizing $U_m$ is $O[ r \prod_{m' \neq m} d_{m'} (r + d_m) + T_2 ( (r d_m N_m' + r^2 d_m) + T_1 (r d_m N_m' + r^2 d_m + r^{2.807} ) ) ]$, where $T_1$ is the number of checks that is needed to find the step size, and $T_2$ is the number of iterations for reaching the stop criterion. Considering that the optimal rank $r \ll d_m$, we can simplify the cost as $O[ r \prod_{m=1}^M d_m + T_2 T_1 ( r d_m N_m' + r^2 d_m ) ]$. Finally, suppose the number of iterations for alternately updating all $\{ U_m \}_{m=1}^M$ is $\Gamma$, we obtain the time complexity of the proposed HMTML, i.e., $¦¯( \Gamma M [ r \prod_{m=1}^M d_m + T_2 T_1 ( r \bar{d}_m \bar{N}_m' + r^2 \bar{d}_m ) ] )$, where $\bar{N}_m'$ and $\bar{d}_m$ are average sample number and feature dimension of all domains respectively. This is linear w.r.t. $M$ and $\prod_{m=1}^M d_m$, and quadratic in the numbers $r$ and $\bar{N}_m$ ($\bar{N}_m'$ is quadratic w.r.t. $\bar{N}_m$). Besides, it is common that $\Gamma < 10$, $T_2 < 20$, and $T_1 < 50$. Thus the complexity is not very high.


\section{Experiments}
\label{sec:Experiments}

In this section, the effectiveness and efficiency of the proposed HMTML are verified with experiments in various applications, i.e., document categorization, scene classification, and image annotation. In the following, we first present the datasets to be used and experimental setups.

\subsection{Datasets and features}
\textbf{Document categorization:} we adopt the Reuters multilingual collection (RMLC) \cite{M-Amini-et-al-NIPS-2009} dataset with six populous categories. The news articles in this dataset are translated into five languages. The TF-IDF features are provided to represent each document. In our experiments, three languages (i.e., English, Spanish, and Italian) are chosen and each language is regarded as one domain. The provided representations are preprocessed by applying principal component analysis (PCA). After the preprocessing, $20\%$ energy is preserved. The resulting dimensions of the document representations are $245$, $213$, and $107$ respectively for the three different domains. The data preprocessing is mainly conducted to: 1) find the high-level patterns which are comparable for transfer; 2) avoid overfitting; 3) speed up the computation during the experiments. The main reasons for preserving only $20\%$ energy are: 1) the amount of energy being preserved does not has influence on the comparison of the effectiveness verification between the proposed approach and other methods. 2) it may lead to overfitting when the feature dimension of each domain is high. There are $18758$, $24039$, and $12342$ instances in the three different domains respectively. In each domain, the sizes of the training and test sets are the same.

\textbf{Scene classification:} We employ a popular natural scene dataset: Scene-15 \cite{S-Lazebnik-et-al-CVPR-2006}. The dataset consists of $4,585$ images that belong to $15$ categories. The features we used are the popular global feature GIST \cite{A-Oliva-and-A-Torralba-IJCV-2001}, local binary pattern (LBP) \cite{T-Ojala-et-al-TPAMI-2002}, and pyramid histogram of oriented gradient (PHOG) \cite{A-Bosch-et-al-ICCV-2007}, where the dimensions are $20$, $59$, and $40$ respectively. A detailed description on how to extract these features can be found in \cite{DX-Dai-et-al-CVPR-2015}. Since the images usually lie in a nonlinear feature space, we preprocess the different features using the kernel PCA (KPCA) \cite{B-Scholkopf-et-al-NCn-1998}. The result dimensions are the same as the original features.

\textbf{Image annotation:} A challenging dataset NUS-WIDE (NUS) \cite{TS-Chua-et-al-CIVR-2009}, which consists of $269,648$ natural images is adopted. We conducted experiments on a subset of $16,519$ images with $12$ animal concepts: bear, bird, cat, cow, dog, elk, fish, fox, horse, tiger, whale, and zebra. In this subset, three kinds of features are extracted for each image: $500$-D bag of visual words \cite{S-Lazebnik-et-al-CVPR-2006} based on SIFT descriptors \cite{DG-Lowe-IJCV-2004}, $144$-D color auto-correlogram, and $128$-D wavelet texture. We refer to \cite{TS-Chua-et-al-CIVR-2009} for a detailed description of the features. Similar to the scene classification, KPCA is adopted for preprocess, and the resulting dimensions are all $100$.

In both scene classification and image annotation, we treat each feature space as a domain. In each domain, half of the samples are used for training, and the rest for test.

\subsection{Experimental setup}

The methods included for comparison are:
\begin{itemize}
  \item \textbf{EU:} directly using the simple Euclidean metric and original feature representations to compute the distance between samples.
  \item \textbf{LMNN \cite{KQ-Weinberger-et-al-NIPS-2005}:} the large margin nearest neighbor DML algorithm presented in \cite{KQ-Weinberger-et-al-NIPS-2005}. Distance metric for different domains are learned separately. The number of local target neighbors is chosen from the set $\{ 1, 2, \ldots, 15 \}$.
  \item \textbf{ITML \cite{JV-Davis-et-al-ICML-2007}:} the information-theoretic metric learning algorithm presented in \cite{JV-Davis-et-al-ICML-2007}. The trade-off hyper-parameter is tuned over the set $\{ 10^i | i = -5, -4, \ldots, 4 \}$.
  \item \textbf{RDML \cite{R-Jin-et-al-NIPS-2009}:} an efficient and competitive distance metric learning algorithm presented in \cite{R-Jin-et-al-NIPS-2009}. We tune trade-off hyper-parameter over the set $\{ 10^i | i = -5, -4, \ldots, 4 \}$.
  \item \textbf{DAMA \cite{C-Wang-and-S-Mahadevan-IJCAI-2011}:} constructing mappings ${U_m}$ to link multiple heterogeneous domains using manifold alignment. The hyper-parameter is determined according to the strategy presented in \cite{C-Wang-and-S-Mahadevan-IJCAI-2011}.
  \item \textbf{MTDA \cite{Y-Zhang-and-DY-Yeung-AAAI-2011}:} applying supervised dimension reduction to heterogeneous representations (domains) simultaneously, where a multi-task extension of linear discriminant analysis is developed. The learned transformation $U_m = W_m H$ consists of a domain specific part $W_m \in \mathbb{R}^{d_m \times d'}$, and a common part $H \in \mathbb{R}^{d' \times r}$ shared by all domains. Here, $d' > r$ is the intermediate dimensionality. As presented in \cite{Y-Zhang-and-DY-Yeung-AAAI-2011}, the shared structure $H$ represents ``some characteristics of the application itself in the common lower-dimensional space''. The hyper-parameter $d'$ is set as $100$ since the model is not very sensitive to the hyper-parameter according to \cite{Y-Zhang-and-DY-Yeung-AAAI-2011}.
  \item \textbf{HMTML:} the proposed heterogeneous multi-task metric learning method. We adopt linear SVMs to obtain the weight vectors of base classifiers for knowledge transfer. The hyper-parameters $\{ \gamma_m \}$ are set as the same value since the different features have been normalized. Both $\gamma$ and $\gamma_m$ are optimized over the set $\{ 10^i | i = -5, -4, \ldots, 4 \}$. It is not hard to determine the value of hyper-parameter $P$, which is the number of base classifiers. According to \cite{EL-Allwein-et-al-JMLR-2000, S-Escalera-et-al-PRL-2009}, the code length is suggested to be $15 \log C$ if we use the sparse random design coding technique in ECOC. From the experimental results shown in \cite{JT-Zhou-et-al-AISTATS-2014}, higher classification accuracy can be achieved with an increasing number $P$, and when there are too many base classifiers, the accuracy may decrease. The optimal $P$ is achieved around (slightly larger than) $15 \log C$, we thus empirically set $P = 10 \lceil 1.5 \log C \rceil$ in our method.
\end{itemize}

The single domain metric learning algorithms (LMNN, ITML, and RDML) only utilize the limited labeled samples in each domain, and do not make use of any additional information from other domains. In DAMA and MTDA, after learning $U_m$, we derive the metric for each domain as $A_m = U_m U_m^T$.


The task in each domain is to perform multi-class classification. The penalty hyper-parameters of the base SVM classifiers are empirically set to $1$. For all compared methods, any types of classifiers can be utilized for final classification in each domain after learning the distance metrics for all domains. In the distance metric learning literature, it is common to use the nearest neighbour ($1$NN) classifier to directly evaluate the performance of the learned metric \cite{ZJ-Zha-et-al-IJCAI-2009, Y-Zhang-and-DY-Yeung-TIST-2012, GJ-Qi-et-al-SDM-2012}, so we mainly adopt it in this paper. Some results of using the sophisticated SVM classifiers can be found in the supplementary material.

In the KPCA preprocessing, we adopt the Gaussian kernel, i.e., $k(\mathbf{x}_i, \mathbf{x}_j) = \mathrm{exp}(-\|\mathbf{x}_i - \mathbf{x}_j\|^2 / (2\omega^2))$, where the hyper-parameter $\omega$ is empirically set as the mean of the distances between all training sample pairs.

In all experiments given below, we adopt the classification accuracy and macroF1 \cite{M-Sokolova-and-G-Lapalme-JIPM-2009} score to evaluate the performance. The average performance of all domains is calculated for comparison. The experiments are run for ten times by randomly choosing different sets of labeled instances. Both the mean values and standard deviations are reported.

If unspecified, the hyper-parameters are determined using leave-one-out cross validation on the labeled set. For example, when the number of labeled samples is $5$ for each class in each domain, one sample is chosen for model evaluation and the remained four samples are used for model training. This leads to $C$ test examples and $4C$ training examples in each domain, where $C$ is the number of classes. This procedure is repeated five times by regarding one of the five labeled samples as test example in turn. The average classification accuracies of all domains from all the five runs are averaged for hyper-parameter determination. The chosen of an optimal common factors of multiple heterogeneous features is still an open problem \cite{DP-Foster-et-al-TR-TTI-2008}, and we do not study it in this paper. To this end, for DAMA, MTDA, and the proposed HMTML, we do not tune the hyper-parameter $r$, which is the number of common factors (or dimensionality of the common subspace) used to explain the original data of all domains. The performance comparisons are performed on a set of varied $r = \{ 1,2,5,8,10,20,30,50,80,100 \}$. The maximum value of $r$ is $20$ in scene classification since the lowest feature dimension is $20$.

%
%

\begin{figure*}
\centering
\subfigure{\includegraphics[width=0.6\columnwidth]{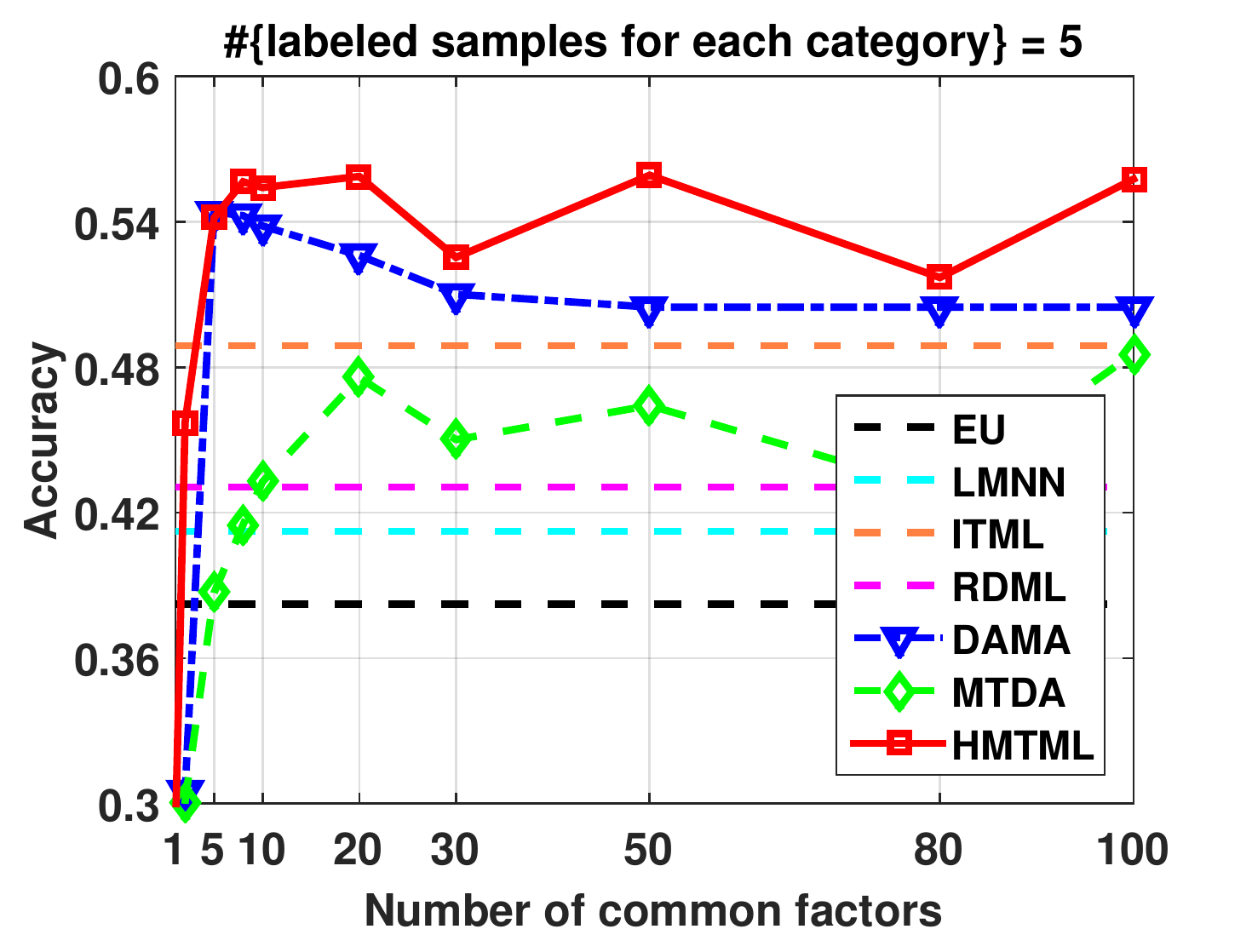}
}
\hfil
\subfigure{\includegraphics[width=0.6\columnwidth]{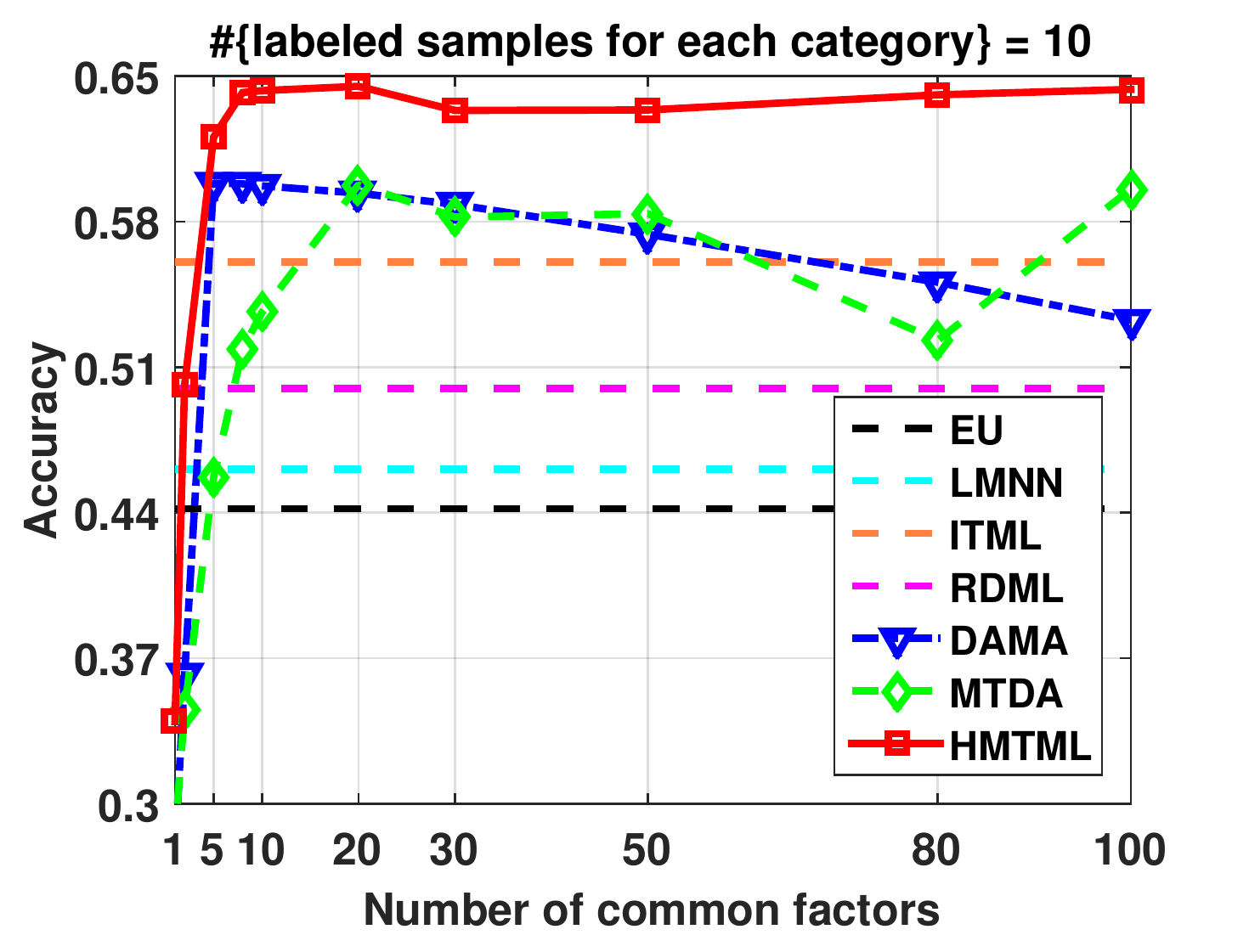}
}
\hfil
\subfigure{\includegraphics[width=0.6\columnwidth]{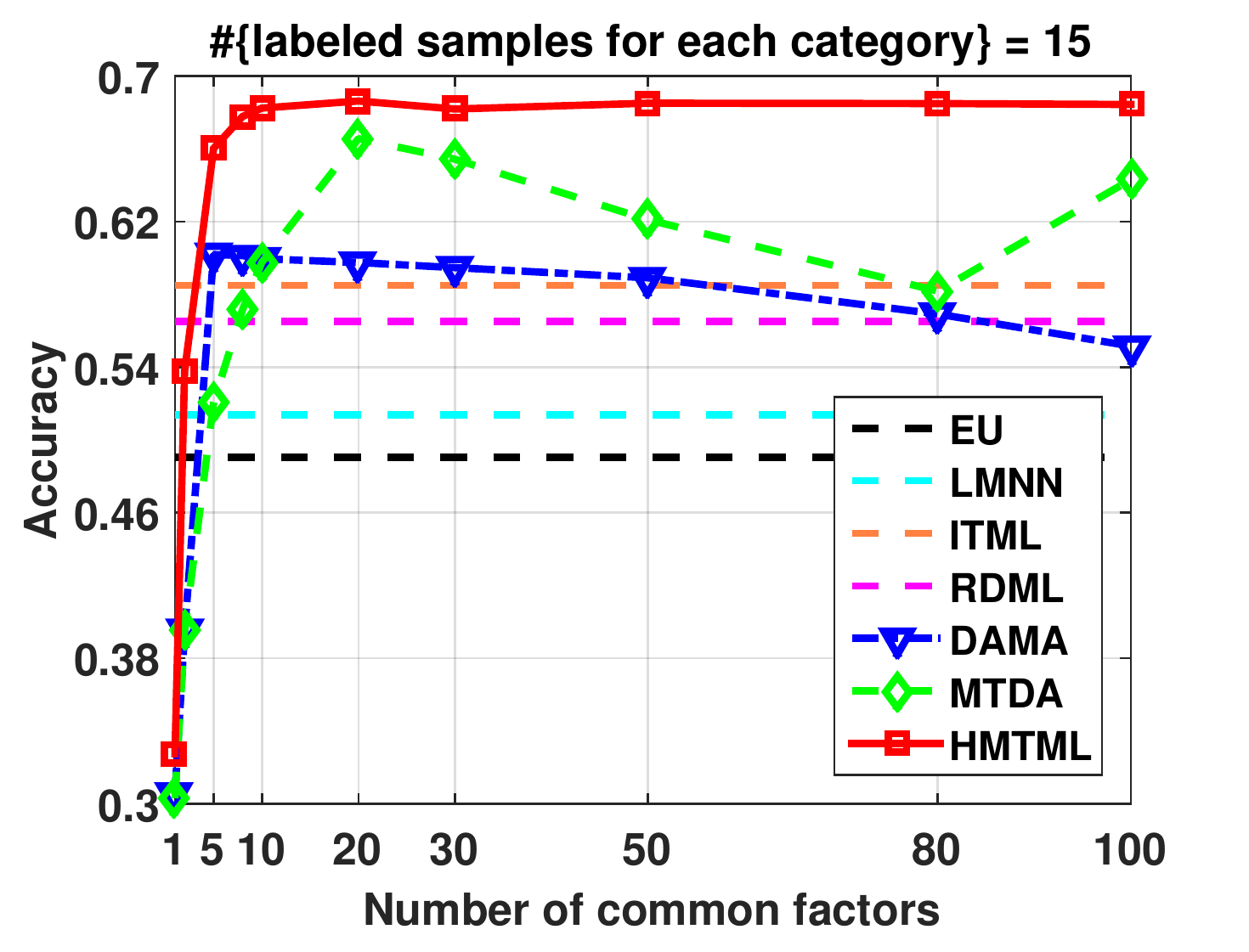}
}
\caption{Average accuracy of all domains vs. number of the common factors on the RMLC dataset.}
\label{fig:Acc_vs_Dim_RMLC}
\end{figure*}

\begin{figure*}
\centering
\subfigure{\includegraphics[width=0.6\columnwidth]{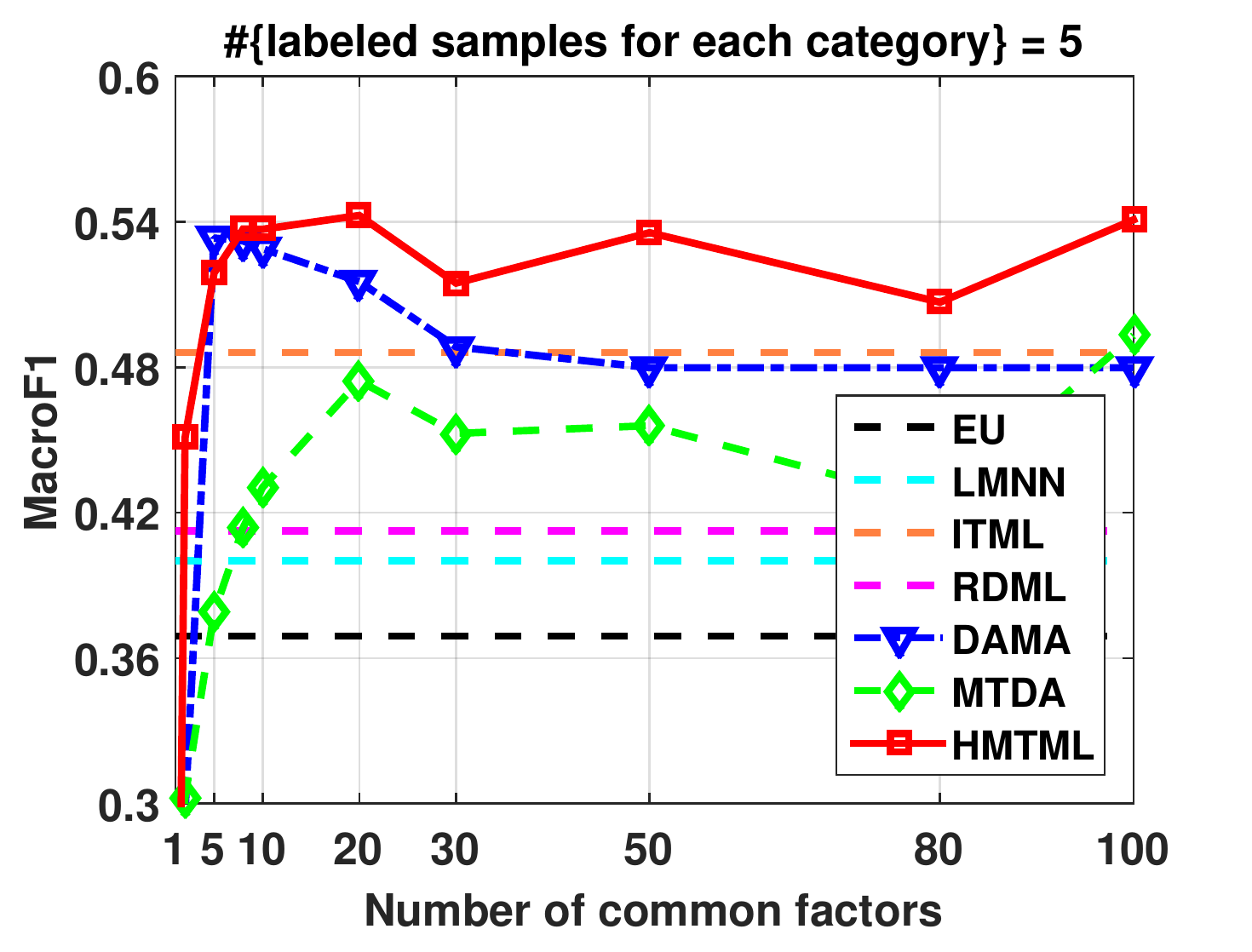}
}
\hfil
\subfigure{\includegraphics[width=0.6\columnwidth]{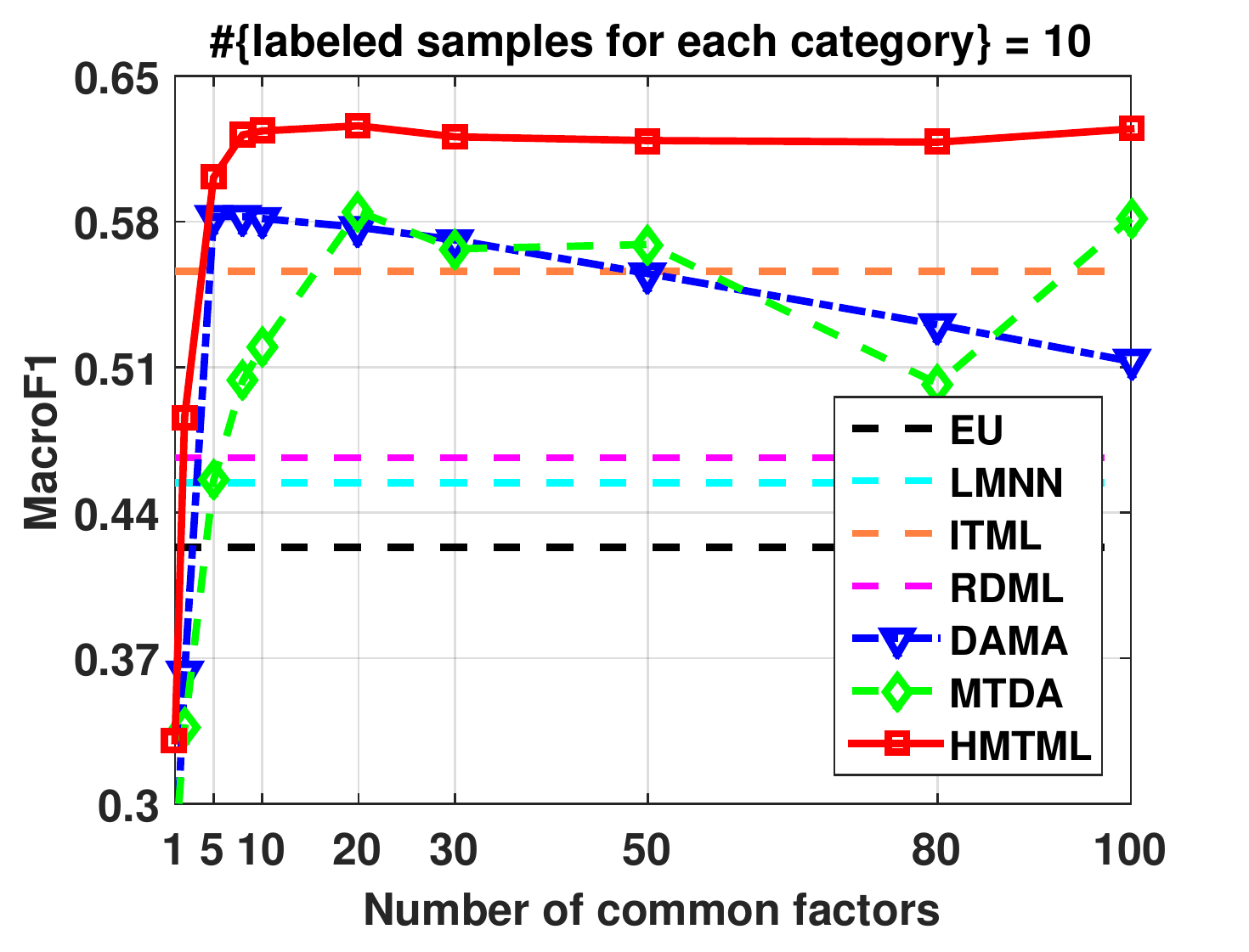}
}
\hfil
\subfigure{\includegraphics[width=0.6\columnwidth]{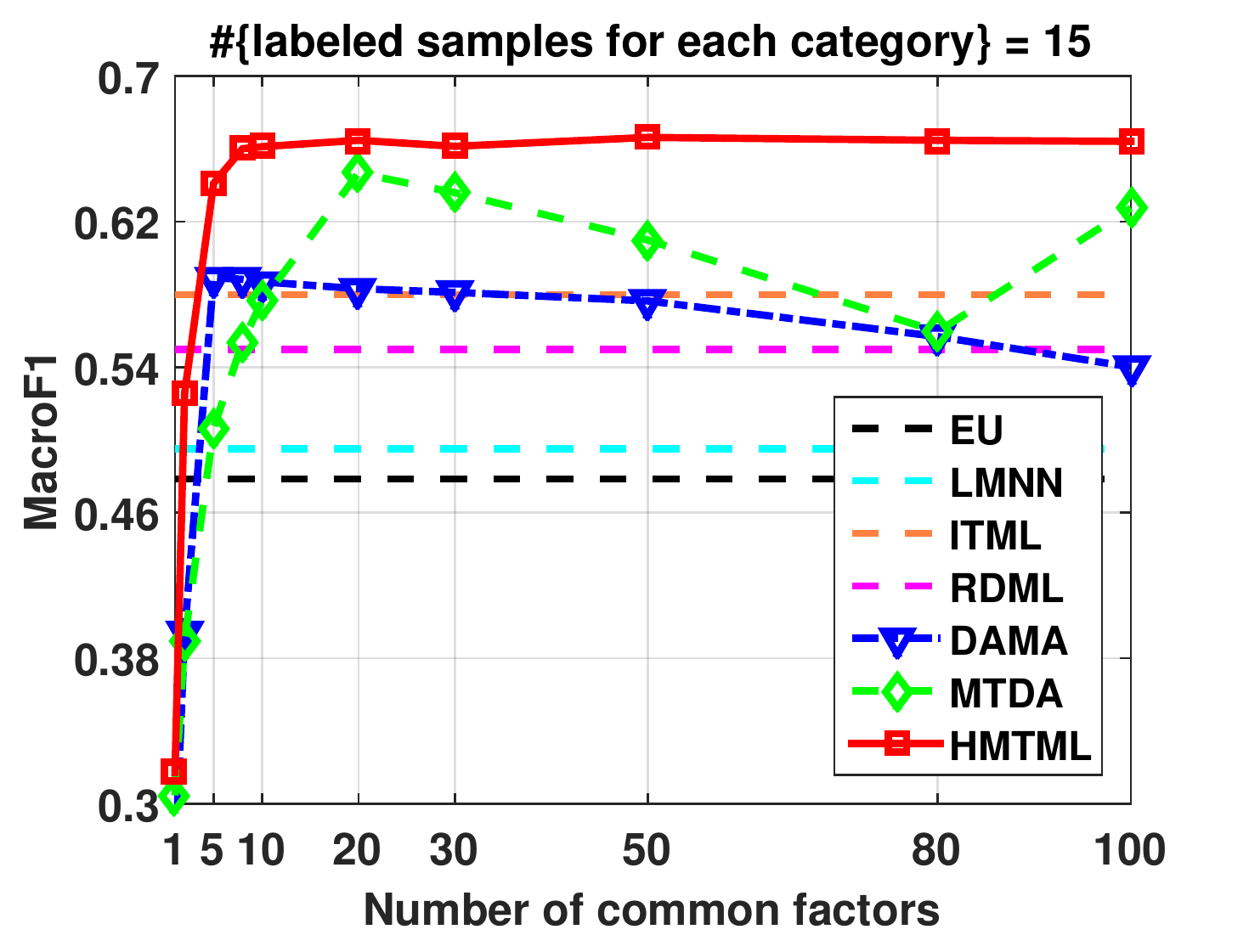}
}
\caption{Average macroF1 score of all domains vs. number of the common factors on the RMLC dataset.}
\label{fig:MacroF1_vs_Dim_RMLC}
\end{figure*}

\begin{table*}[!t]
\setlength\tabcolsep{2pt}
\renewcommand{\arraystretch}{1.3}
\caption{Average accuracy and macroF1 score of all domains of the compared methods at their best numbers (of common factors) on the RMLC dataset. In each domain, the number of labeled training samples for each category varies from $5$ to $15$.}
\label{tab:Acc_MacroF1_RMLC}
\centering
\begin{tabular}{c||c|c|c||c|c|c}
\hline
\ & \multicolumn{3}{c||}{Accuracy} & \multicolumn{3}{c}{MacroF1} \\
\hline
Methods & 5 & 10 & 15 & 5 & 10 & 15 \\
\hline
EU & 0.382$\pm$0.023 & 0.442$\pm$0.012 & 0.491$\pm$0.012 & 0.369$\pm$0.030 & 0.423$\pm$0.012 & 0.479$\pm$0.022 \\
\hline
LMNN & 0.412$\pm$0.020 & 0.461$\pm$0.017 & 0.514$\pm$0.017 & 0.400$\pm$0.029 & 0.454$\pm$0.021 & 0.495$\pm$0.015 \\
ITML & 0.489$\pm$0.038 & 0.561$\pm$0.032 & 0.585$\pm$0.026 & 0.486$\pm$0.028 & 0.556$\pm$0.024 & 0.580$\pm$0.024 \\
RDML & 0.431$\pm$0.014 & 0.500$\pm$0.029 & 0.565$\pm$0.013 & 0.412$\pm$0.019 & 0.466$\pm$0.023 & 0.550$\pm$0.017 \\
\hline
DAMA & 0.544$\pm$0.021 & 0.599$\pm$0.029 & 0.602$\pm$0.009 & 0.533$\pm$0.013 & 0.583$\pm$0.027 & 0.589$\pm$0.012 \\
MTDA & 0.486$\pm$0.031 & 0.598$\pm$0.015 & 0.666$\pm$0.008 & 0.493$\pm$0.023 & 0.585$\pm$0.016 & 0.647$\pm$0.012 \\
HMTML & \textbf{0.559$\pm$0.005} & \textbf{0.645$\pm$0.013} & \textbf{0.686$\pm$0.003} & \textbf{0.543$\pm$0.015} & \textbf{0.626$\pm$0.010} & \textbf{0.666$\pm$0.007} \\
\hline
\end{tabular}
\end{table*}

\subsection{Overall classification results}

\textbf{Document categorization:} In the training set, the number of labeled instances for each category is chosen randomly as $\{ 5, 10, 15 \}$. This is used to evaluate the performance of all compared approaches w.r.t. the number of labeled samples. The results of selecting more labeled samples are given in supplementary material. Fig. \ref{fig:Acc_vs_Dim_RMLC} and Fig. \ref{fig:MacroF1_vs_Dim_RMLC} shows the accuracies and macroF1 values respectively for different $r$. The performance of all compared approaches at its best number (of common factors) can be found in Table \ref{tab:Acc_MacroF1_RMLC}. From these results, we can draw several conclusions: 1) when the number of labeled samples increases, the performance of all approaches become better; 2) although limited number of labeled data is given in each domain, ITML and RDML can greatly improve the performance. This indicates that distance metric learning (DML) is an effective tool in this application; 3) the performance of all heterogeneous transfer learning methods (DAMA, MTDA, and HMTML) is much better than the single-domain DML algorithms (LMNN, ITML, and RDML). This indicates that it is useful to leverage information from other domains in DML; In addition, it can be seen from the results that the optimal number $r$ is often smaller than $30$. Hence we may only need $30$ factors to distinguish the different categories in this dataset; 4) although effective and discriminative, the performance of MTDA is heavily dependent on label information. Therefore, when the labeled samples are scarce, MTDA is worse than DAMA. The topology preserving of DAMA is helpful when the labeled samples are insufficient; 5) the proposed HMTML also relies a lot on the label information to learn base classifiers, which are used to connect different domains. If the labeled samples are scarce, some learned classifiers may be unreliable. However, owing to the task construction strategy presented in \cite{JT-Zhou-et-al-AISTATS-2014}, robust transformations can be obtained even we have learnt some incorrect or inaccurate classifiers. Therefore, our method outperforms DAMA even when the labeled instances are scarce; 6) overall, the developed HMTML is superior to both DAMA and MTDA at most numbers (of common factors). This can be interpreted as the expressive ability of the factors learned by our method are stronger than other compared methods. The main reason is that the high-order statistics of all domains are examined. However, only the pairwise correlations are exploited in DAMA and MTDA; 7) consistent performance have been achieved under different criteria (accuracy and macroF1 score). Specifically, compared with the competitive MTDA, the improvements of the proposed method are considerable. For example, when $5$, $10$, and $15$ labeled samples are utilized, the relative improvements are $10.0\%$, $7.1\%$, and $3.1\%$ respectively under macroF1 criteria.

In addition, we conduct experiments on more domains to verify that the proposed method is able to handle arbitrary number of heterogeneous domains. See the supplementary material for the results.

\begin{figure*}
\centering
\subfigure{\includegraphics[width=0.6\columnwidth]{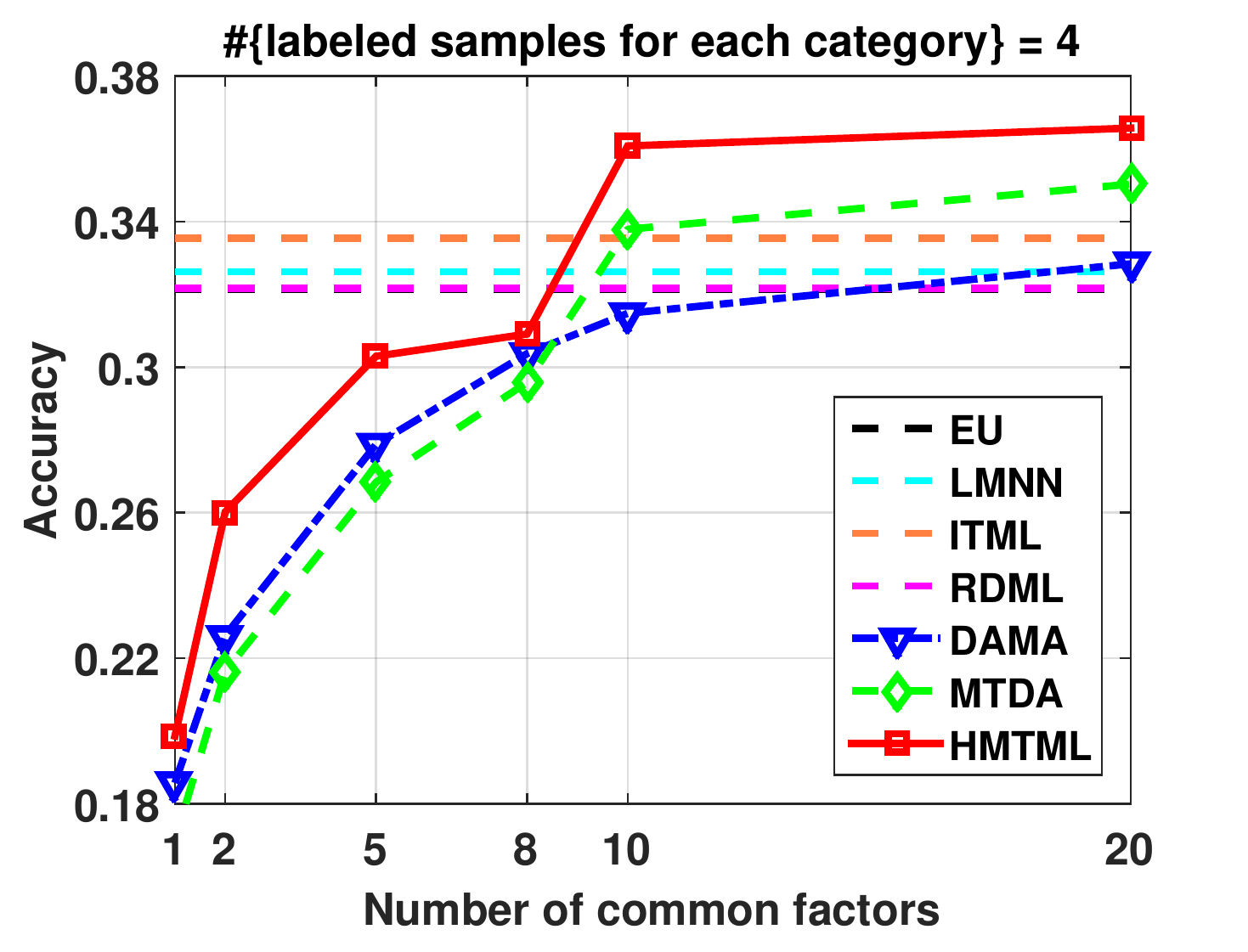}
}
\hfil
\subfigure{\includegraphics[width=0.6\columnwidth]{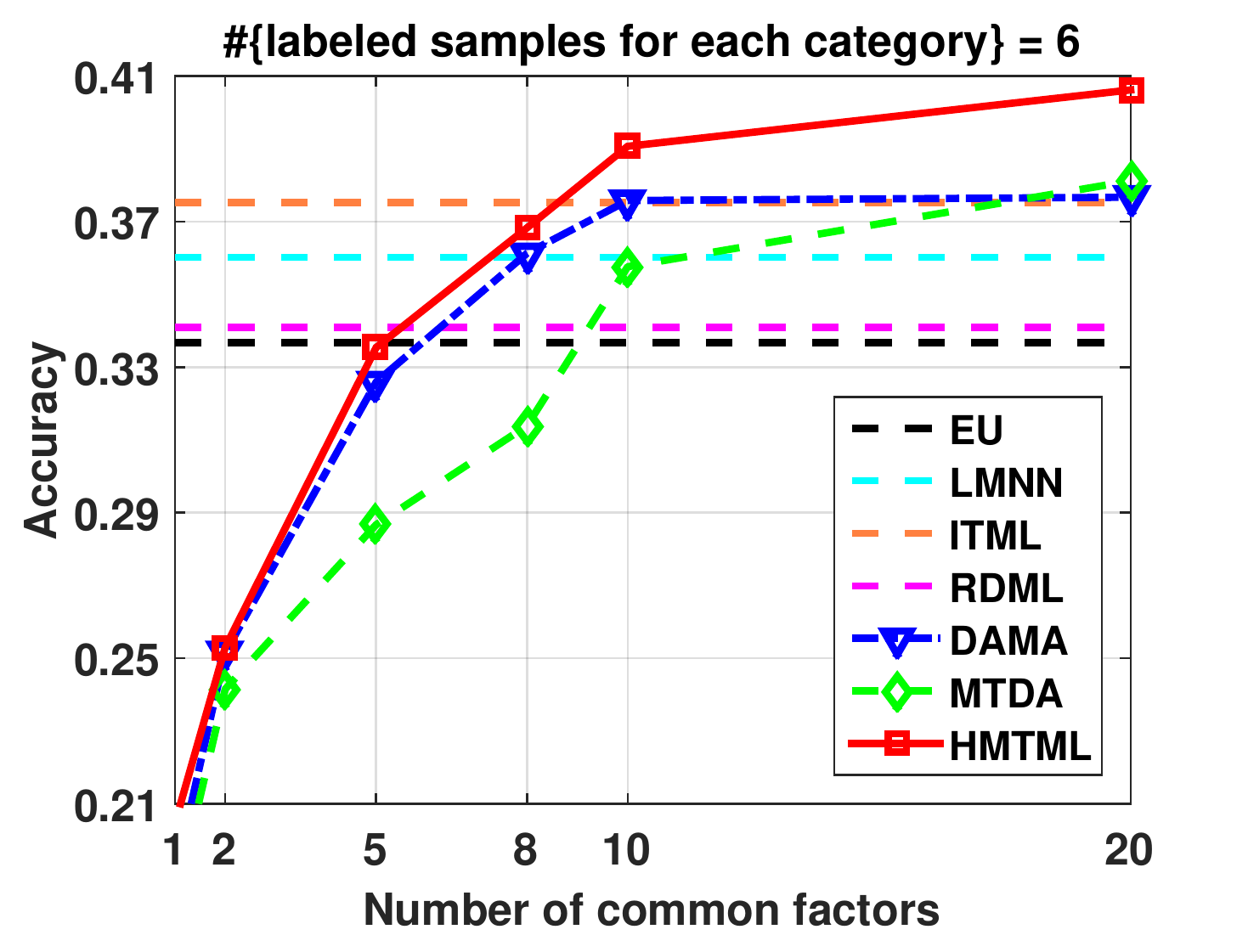}
}
\hfil
\subfigure{\includegraphics[width=0.6\columnwidth]{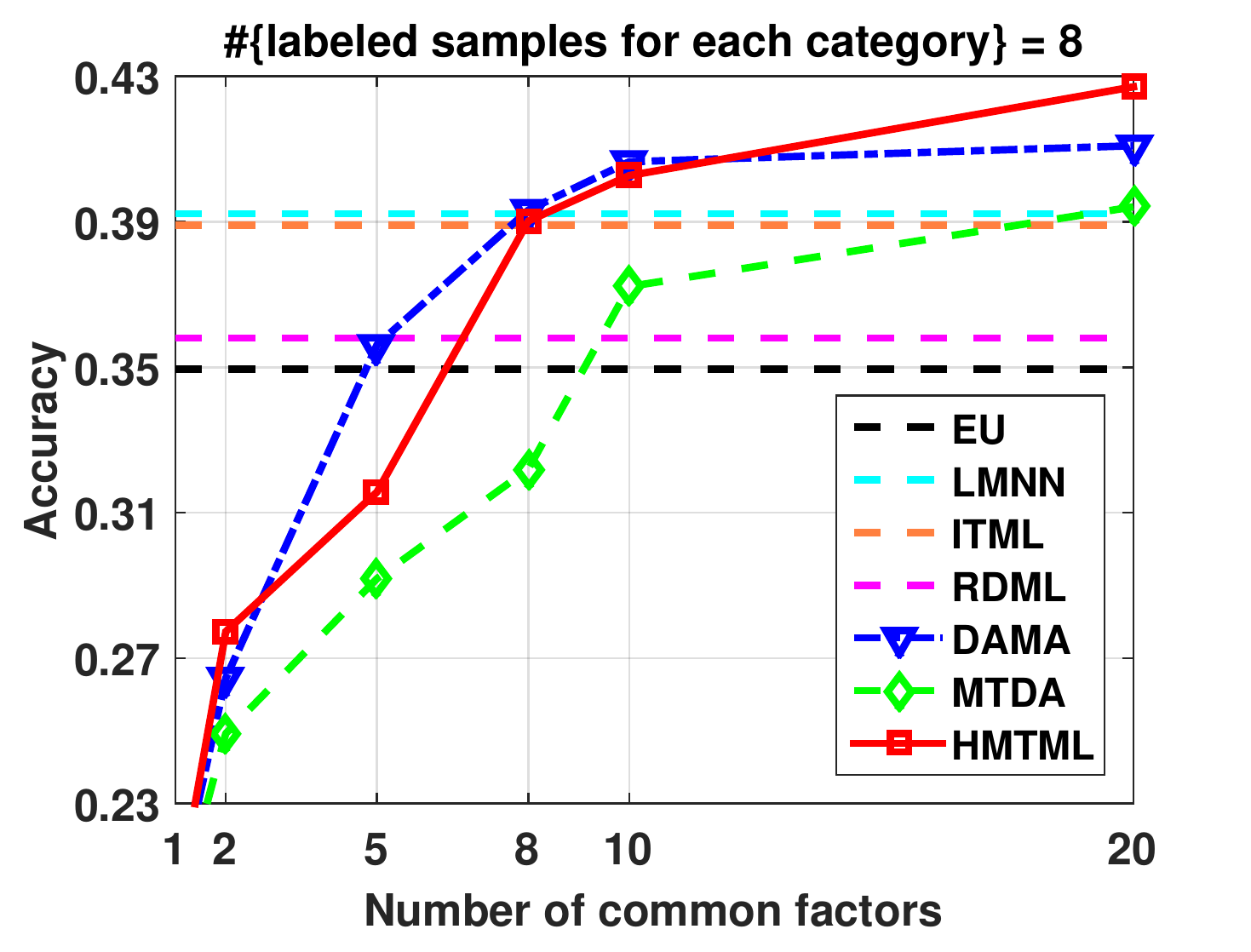}
}
\caption{Average accuracy of all domains vs. number of the common factors on the Scene-15 dataset.}
\label{fig:Acc_vs_Dim_SCE}
\end{figure*}

\begin{figure*}
\centering
\subfigure{\includegraphics[width=0.6\columnwidth]{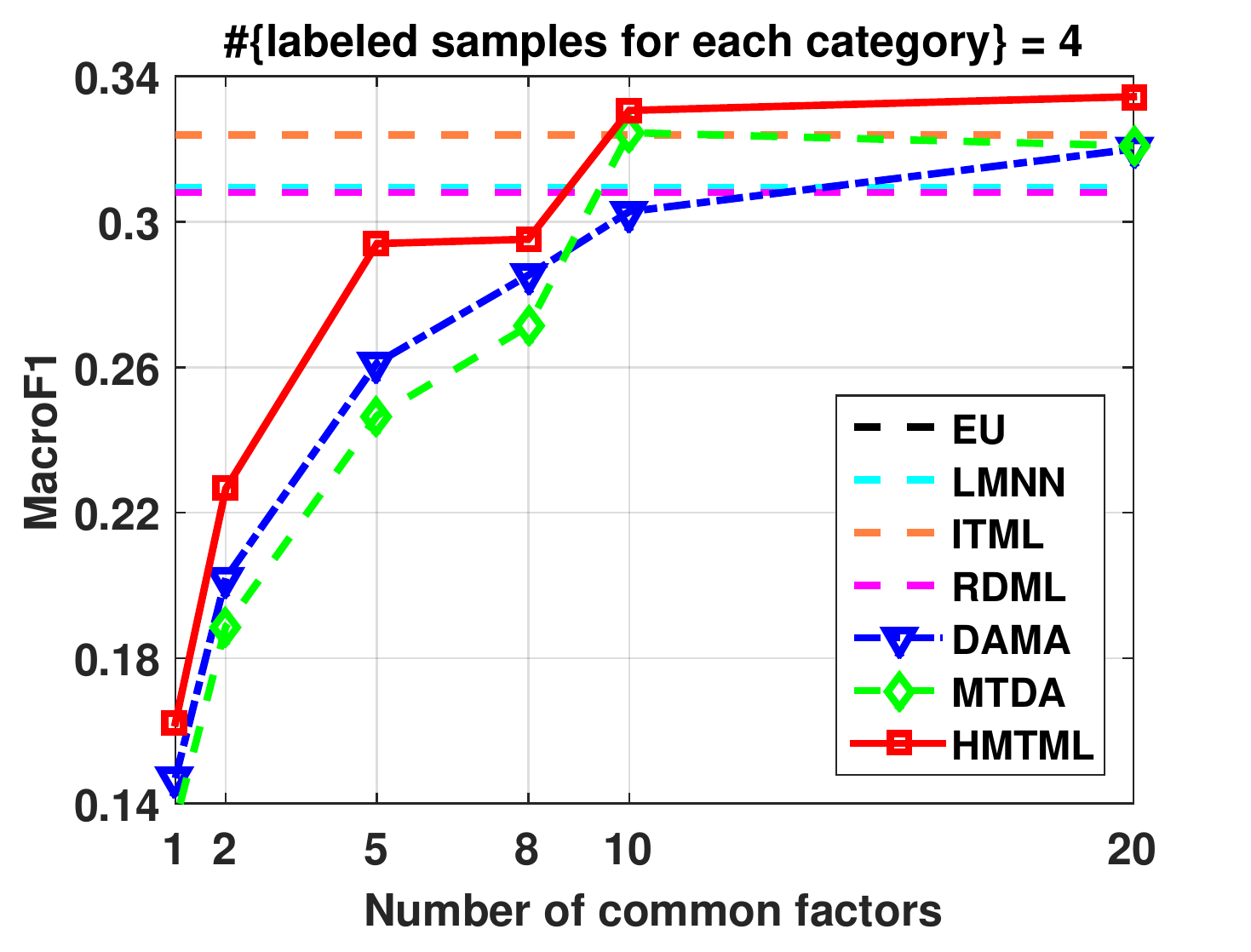}
}
\hfil
\subfigure{\includegraphics[width=0.6\columnwidth]{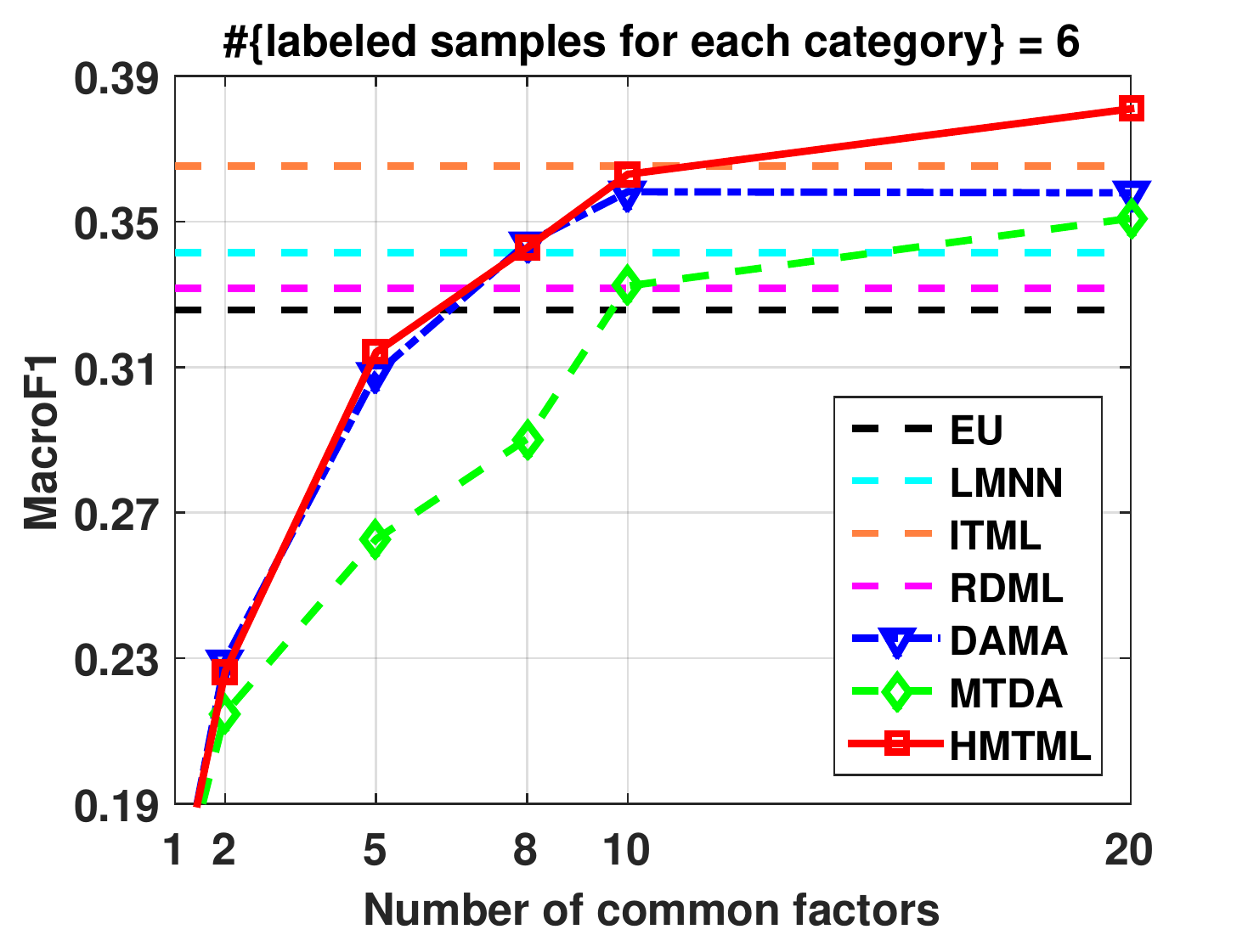}
}
\hfil
\subfigure{\includegraphics[width=0.6\columnwidth]{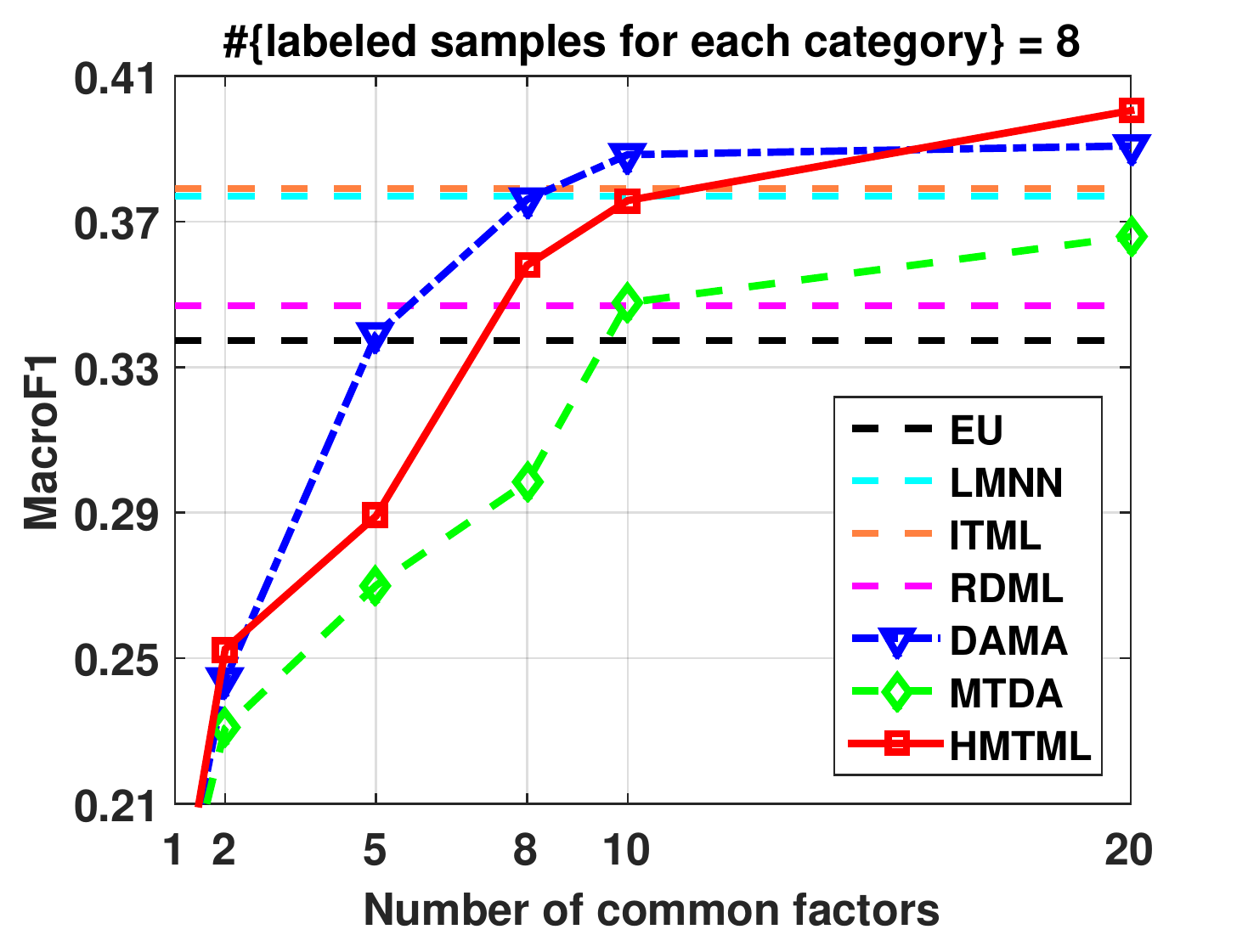}
}
\caption{Average macroF1 score of all domains vs. number of the common factors on the Scene-15 dataset.}
\label{fig:MacroF1_vs_Dim_SCE}
\end{figure*}

\begin{table*}[!t]
\setlength\tabcolsep{3pt}
\renewcommand{\arraystretch}{1.3}
\caption{Average accuracy and macroF1 score of all domains of the compared methods at their best numbers (of common factors) on the Scene-15 dataset. In each domain, the number of labeled training examples for each category varies from $4$ to $8$.}
\label{tab:Acc_MacroF1_SCE}
\centering
\begin{tabular}{c||c|c|c||c|c|c}
\hline
\ & \multicolumn{3}{c||}{Accuracy} & \multicolumn{3}{c}{MacroF1} \\
\hline
Methods & 4 & 6 & 8 & 4 & 6 & 8 \\
\hline
EU & 0.321$\pm$0.018 & 0.337$\pm$0.012 & 0.349$\pm$0.010 & 0.309$\pm$0.017 & 0.326$\pm$0.012 & 0.337$\pm$0.011 \\
\hline
LMNN & 0.326$\pm$0.014 & 0.360$\pm$0.006 & 0.392$\pm$0.011 & 0.310$\pm$0.010 & 0.342$\pm$0.005 & 0.377$\pm$0.011 \\
ITML & 0.336$\pm$0.017 & 0.375$\pm$0.009 & 0.389$\pm$0.011 & 0.324$\pm$0.017 & 0.365$\pm$0.009 & 0.379$\pm$0.011 \\
RDML & 0.322$\pm$0.005 & 0.341$\pm$0.004 & 0.358$\pm$0.005 & 0.308$\pm$0.009 & 0.332$\pm$0.008 & 0.347$\pm$0.007 \\
\hline
DAMA & 0.328$\pm$0.009 & 0.377$\pm$0.016 & 0.411$\pm$0.010 & 0.320$\pm$0.007 & 0.358$\pm$0.012 & 0.391$\pm$0.009 \\
MTDA & 0.350$\pm$0.008 & 0.381$\pm$0.006 & 0.394$\pm$0.008 & 0.325$\pm$0.011 & 0.351$\pm$0.006 & 0.366$\pm$0.008 \\
HMTML & \textbf{0.366$\pm$0.008} & \textbf{0.406$\pm$0.011} & \textbf{0.427$\pm$0.004} & \textbf{0.334$\pm$0.009} & \textbf{0.381$\pm$0.012} & \textbf{0.401$\pm$0.004} \\
\hline
\end{tabular}
\end{table*}

\textbf{Scene classification:} The number of labeled examples for each category varies in the set $\{ 4, 6, 8 \}$. The performance w.r.t. the number $r$ are shown in Fig. \ref{fig:Acc_vs_Dim_SCE} and Fig. \ref{fig:MacroF1_vs_Dim_SCE}. The average performance are summarized in Table \ref{tab:Acc_MacroF1_SCE}. We can see from the results that: 1) when the number of labeled examples is $4$, both LMNN and RDML are only comparable to directly using the Euclidean distance (EU). ITML and DAMA are only slightly better than the EU baseline. This may be because the learned metrics are linear, while the images are usually lie in a nonlinear feature space. Although the KPCA preprocessing can help to exploit the nonlinearity to some extent, it does not optimize w.r.t. the metric; 2) when comparing with the EU baseline, the heterogeneous transfer learning approaches do not improve that much as in document categorization. This can be interpreted as that the different domains correspond to different types of representations in this problem of scene classification. It is much more challenging than the classification of multilingual documents, where the same kind of feature (TF-IDF) with various vocabularies is adopted. In this application, the statistical properties of the various types of visual representations greatly differ from each other. Therefore, the common factors contained among these different domains are difficult to be discovered by only exploring the pair-wise correlations between them. However, much better performance can be obtained by the presented HMTML since the correlations of all domains are exploited simultaneously, especially when small number of labeled samples are given. Thus the superiority of our method in alleviating the labeled data deficiency issue is verified.

\begin{figure*}
\centering
\subfigure{\includegraphics[width=0.6\columnwidth]{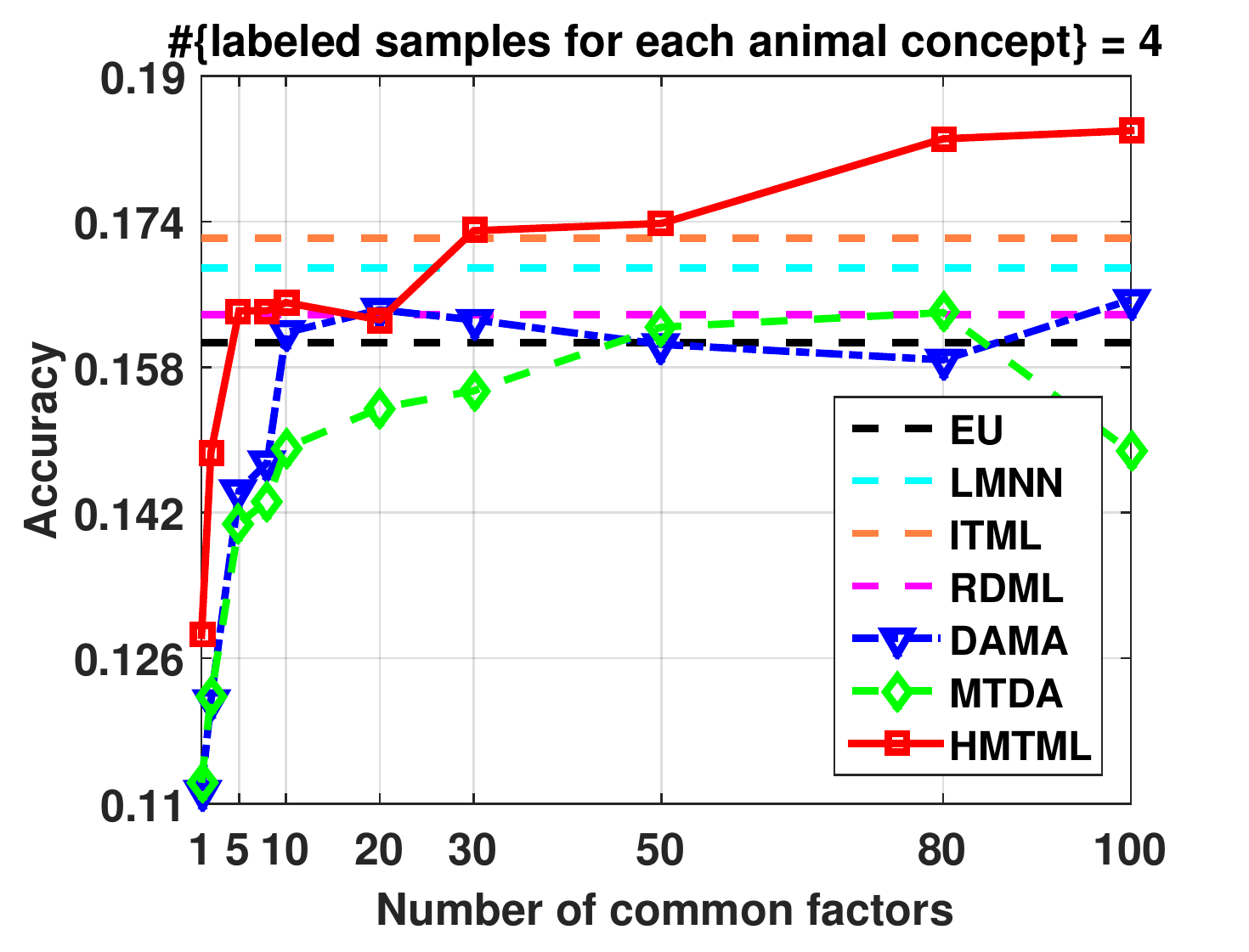}
}
\hfil
\subfigure{\includegraphics[width=0.6\columnwidth]{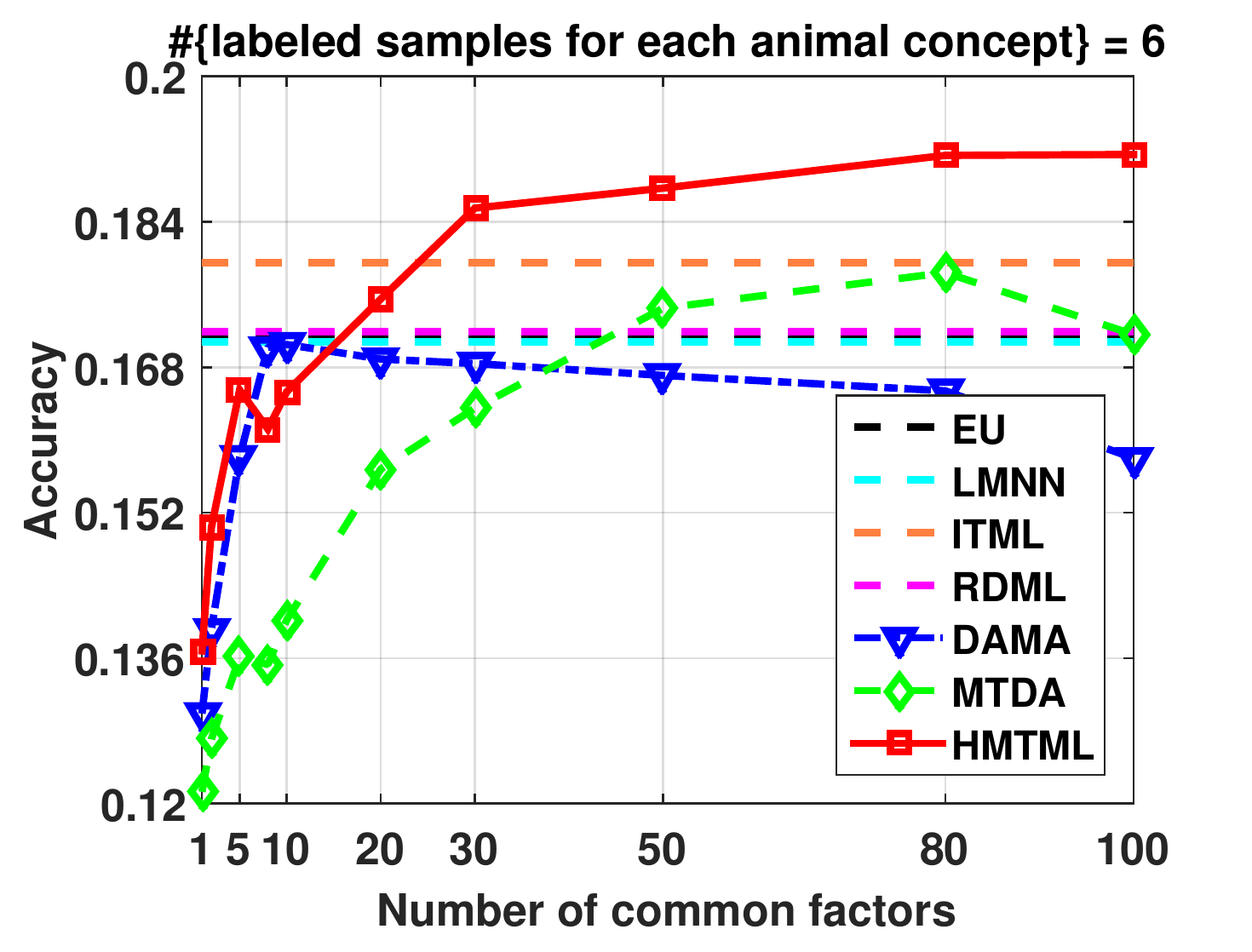}
}
\hfil
\subfigure{\includegraphics[width=0.6\columnwidth]{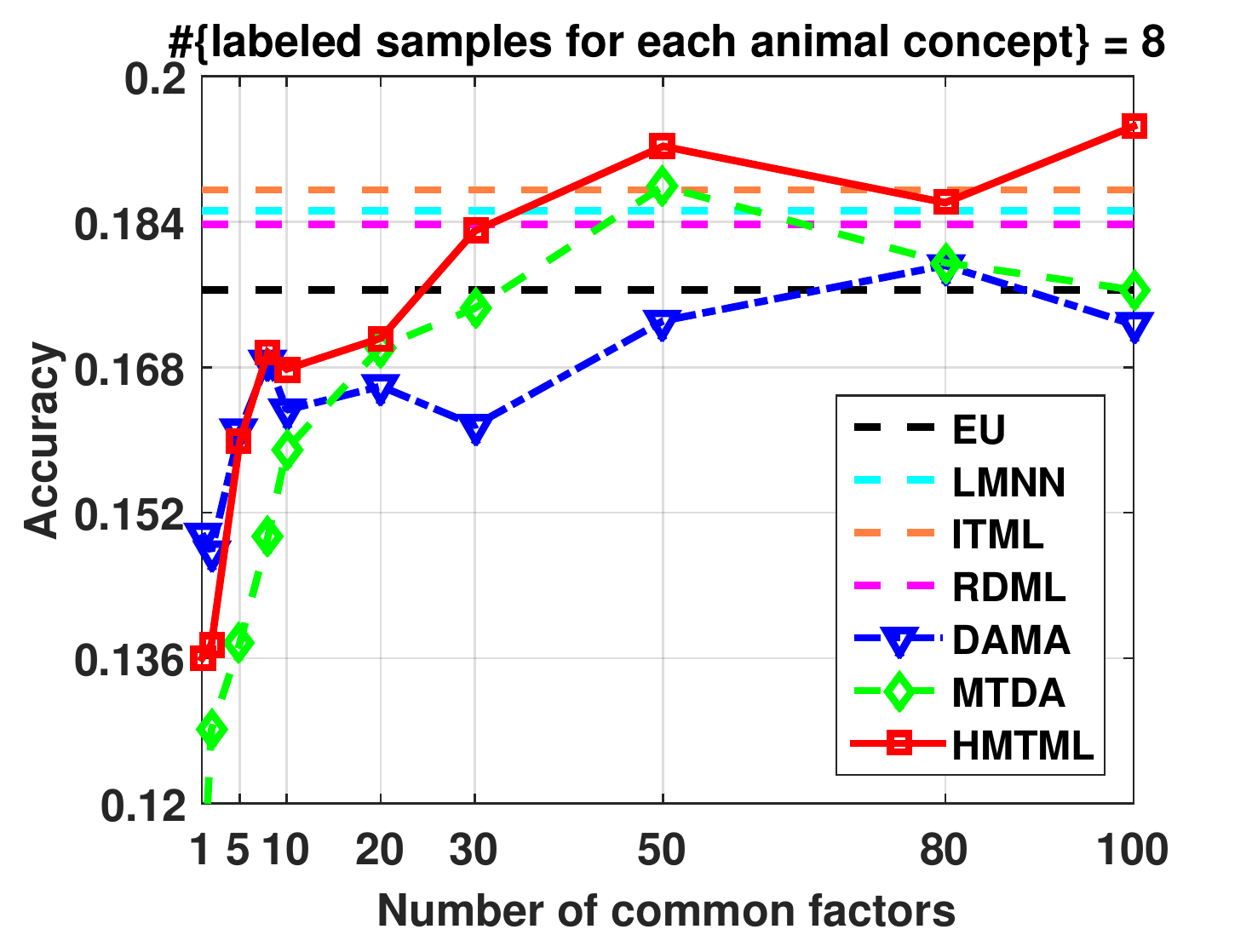}
}
\caption{Average accuracy of all domains vs. number of the common factors on the NUS animal subset.}
\label{fig:Acc_vs_Dim_NUS}
\end{figure*}
\begin{figure*}
\centering
\subfigure{\includegraphics[width=0.6\columnwidth]{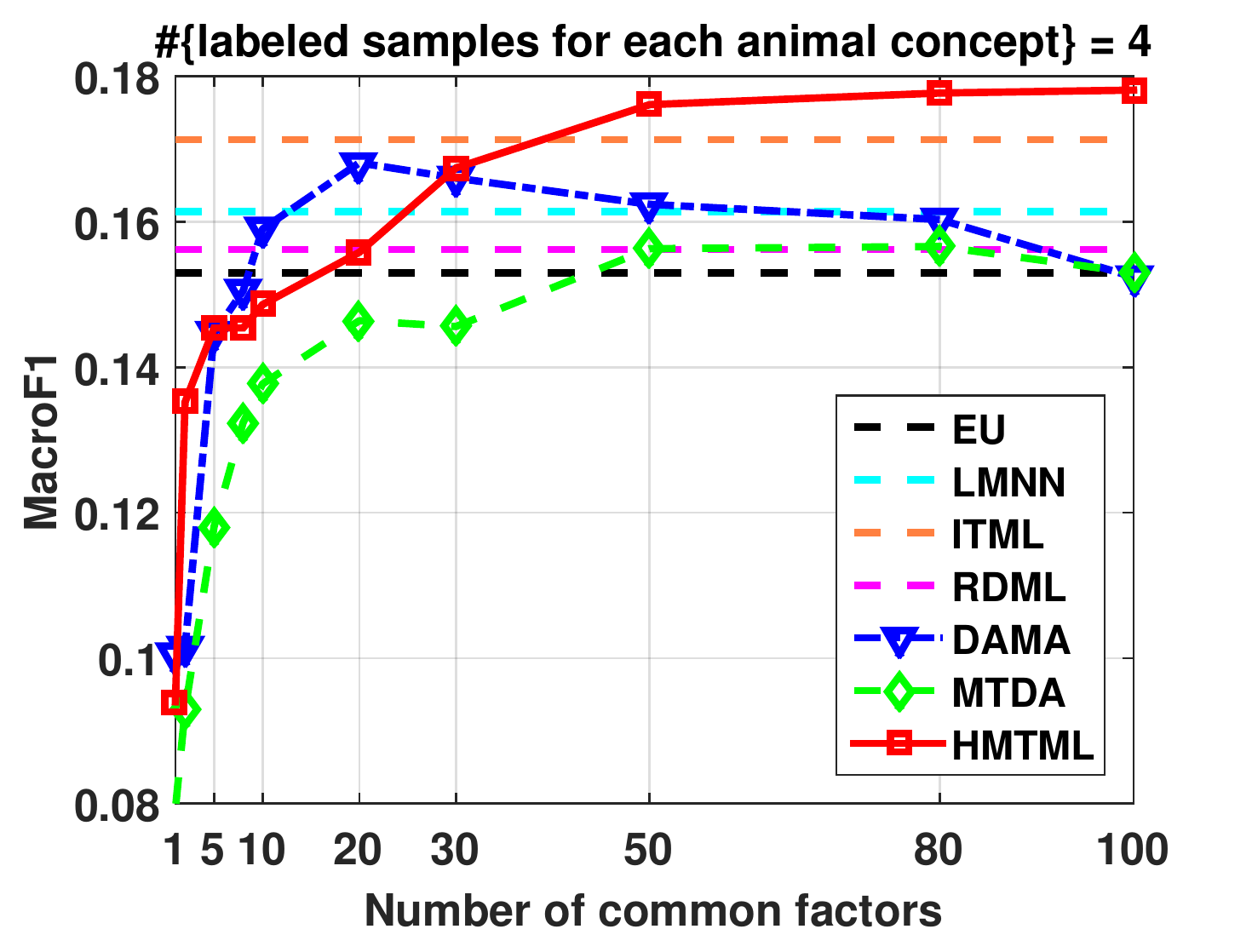}
}
\hfil
\subfigure{\includegraphics[width=0.6\columnwidth]{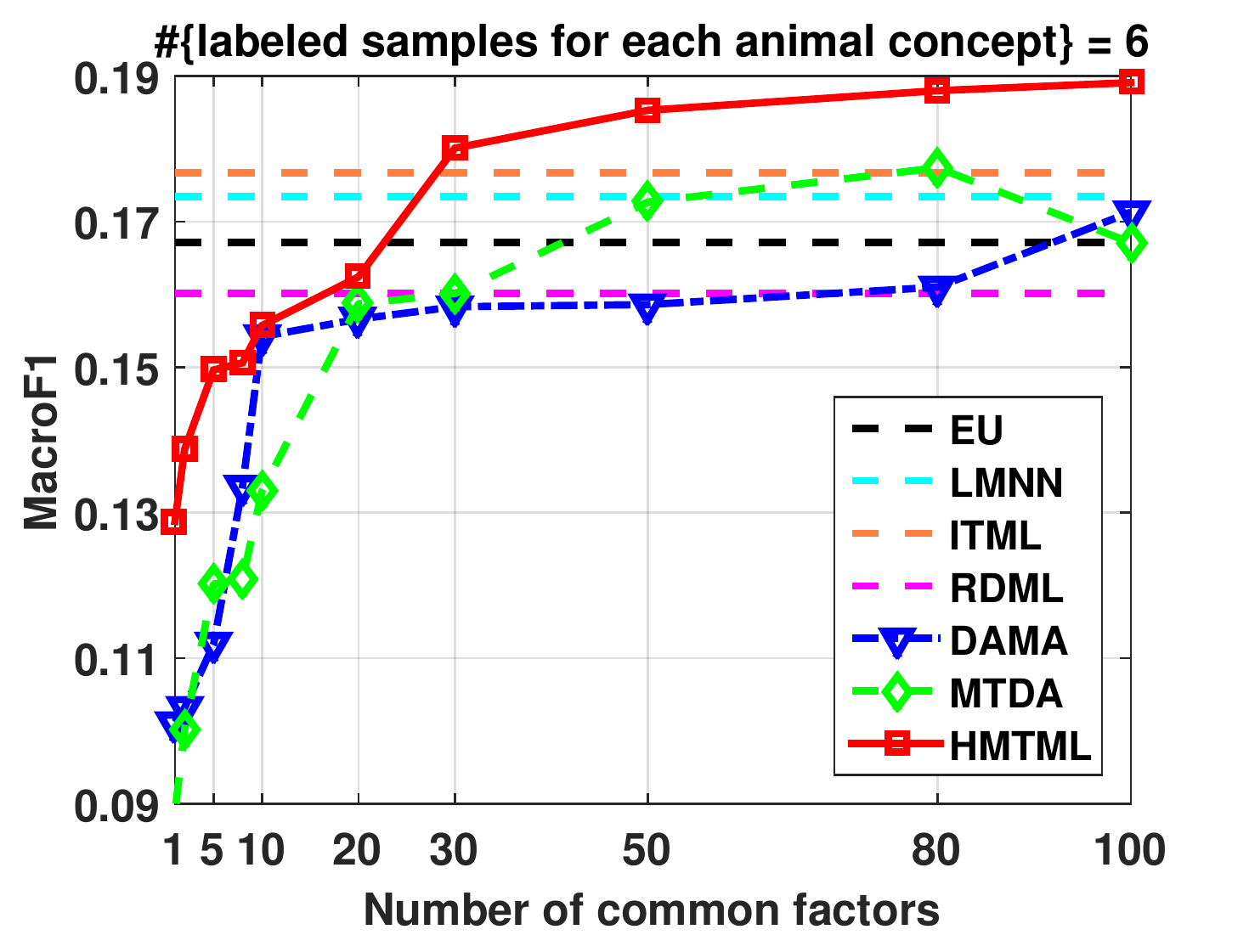}
}
\hfil
\subfigure{\includegraphics[width=0.6\columnwidth]{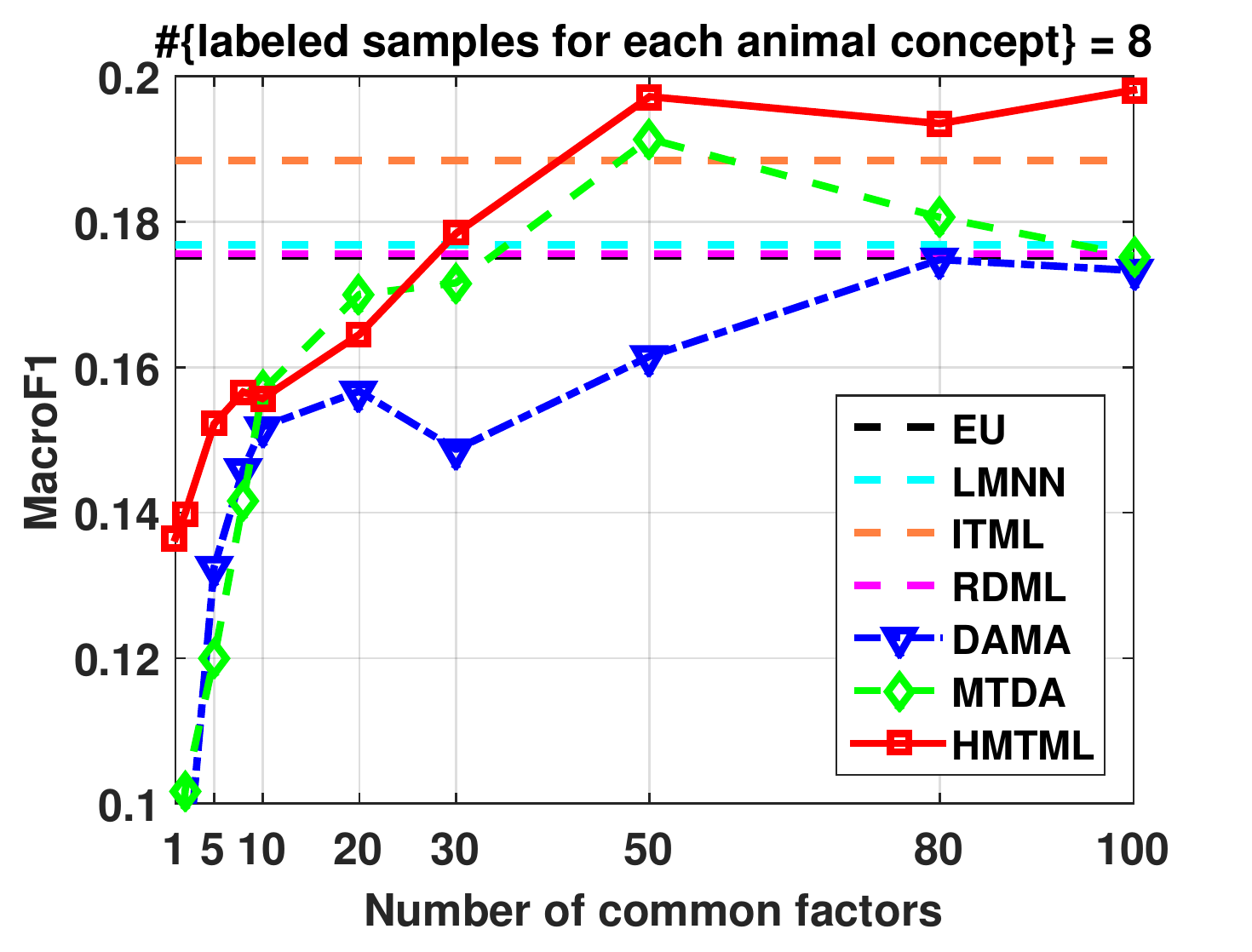}
}
\caption{Average macroF1 score of all domains vs. number of the common factors on the NUS animal subset.}
\label{fig:MacroF1_vs_Dim_NUS}
\end{figure*}

\textbf{Image annotation:} The number of labeled instances for each concept varies in the set $\{ 4, 6, 8 \}$. We show the annotation accuracies and MacroF1 scores of the compared methods in Fig. \ref{fig:Acc_vs_Dim_NUS} and Fig. \ref{fig:MacroF1_vs_Dim_NUS} respectively, and summarize the results at their best numbers (of common factors) in Table \ref{tab:Acc_MacroF1_NUS}. It can be observed from the results that: 1) all of the single domain metric learning algorithms do not perform well. This is mainly due to the challenging of the setting that each kind of feature is regarded as a domain. Besides, in this application, the different animal concepts are hard to distinguish since they have high inter-class similarity (e.g., ``cat'' is similar to ``tiger'') or have large intra-class variability (such as ``bird''). More labeled instances are necessary for these algorithms to work, and we refer to the supplementary material for demonstration; 2) DAMA is comparable to the EU baseline, and MTDA only obtains satisfactory accuracies when enough (e.g., $8$) labeled instances are provided. Whereas the proposed HMTML still outperforms other approaches in most cases, and the tendencies of the macroF1 score curves are the same as accuracy.

\begin{table*}[!t]
\setlength\tabcolsep{3pt}
\renewcommand{\arraystretch}{1.3}
\caption{Average accuracy and macroF1 score of all domains of the compared methods at their best numbers (of common factors) on the NUS animal subset. In each domain, the number of labeled training instances for each concept varies from $4$ to $8$.}
\label{tab:Acc_MacroF1_NUS}
\centering
\begin{tabular}{c||c|c|c||c|c|c}
\hline
\ & \multicolumn{3}{c||}{Accuracy} & \multicolumn{3}{c}{MacroF1} \\
\hline
Methods & 4 & 6 & 8 & 4 & 6 & 8 \\
\hline
EU & 0.161$\pm$0.011 & 0.172$\pm$0.008 & 0.177$\pm$0.009 & 0.153$\pm$0.013 & 0.167$\pm$0.009 & 0.175$\pm$0.007 \\
\hline
LMNN & 0.169$\pm$0.006 & 0.176$\pm$0.001 & 0.185$\pm$0.004 & 0.161$\pm$0.008 & 0.173$\pm$0.004 & 0.177$\pm$0.005 \\
ITML & 0.172$\pm$0.012 & 0.180$\pm$0.008 & 0.188$\pm$0.006 & 0.171$\pm$0.012 & 0.177$\pm$0.009 & 0.188$\pm$0.007 \\
RDML & 0.164$\pm$0.009 & 0.172$\pm$0.009 & 0.184$\pm$0.005 & 0.156$\pm$0.005 & 0.160$\pm$0.011 & 0.176$\pm$0.006 \\
\hline
DAMA & 0.165$\pm$0.008 & 0.171$\pm$0.012 & 0.179$\pm$0.003 & 0.168$\pm$0.006 & 0.171$\pm$0.004 & 0.175$\pm$0.005 \\
MTDA & 0.164$\pm$0.019 & 0.178$\pm$0.009 & 0.188$\pm$0.010 & 0.157$\pm$0.022 & 0.177$\pm$0.011 & 0.191$\pm$0.010 \\
HMTML & \textbf{0.184$\pm$0.010} & \textbf{0.191$\pm$0.004} & \textbf{0.195$\pm$0.004} & \textbf{0.178$\pm$0.010} & \textbf{0.189$\pm$0.003} & \textbf{0.198$\pm$0.007} \\
\hline
\end{tabular}
\end{table*}

\subsection{Further empirical study}

In this section, we conduct some further empirical studies on the RMLC dataset. More results can be found in the supplementary material.

\subsubsection{An investigation of the performance for the individual domains}

\begin{figure}
\centering
\subfigure{\includegraphics[width=0.48\columnwidth]{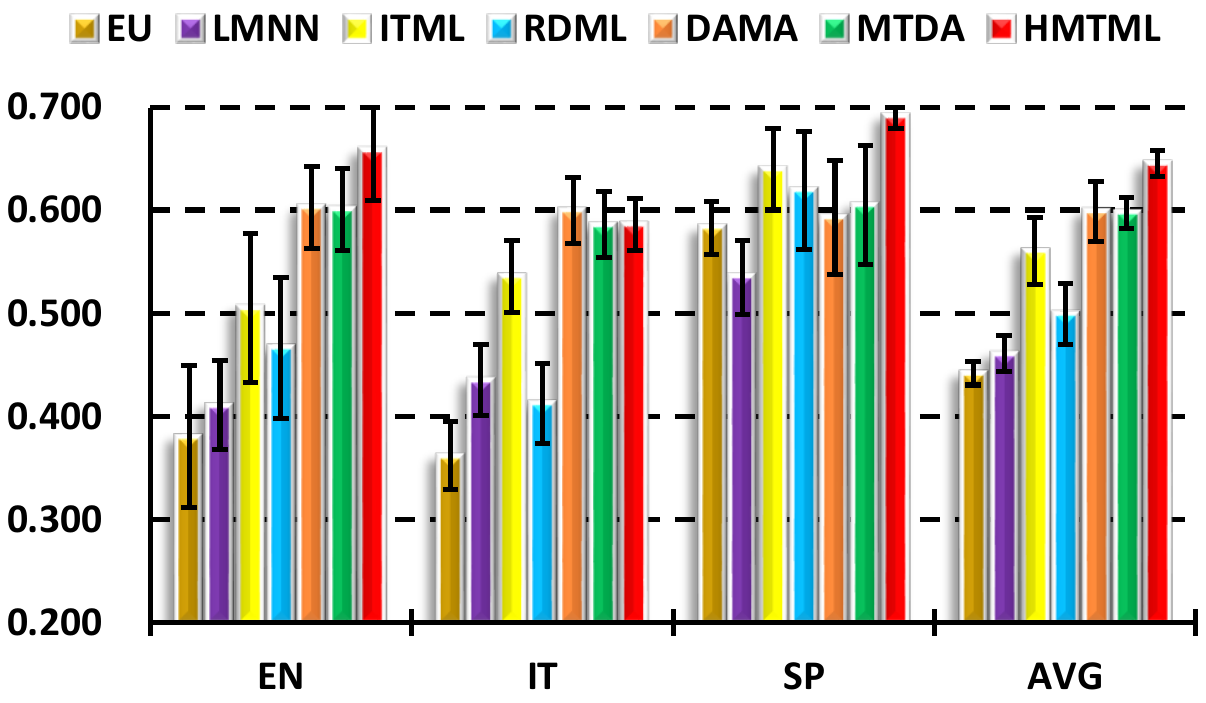}
}
\hfil
\subfigure{\includegraphics[width=0.48\columnwidth]{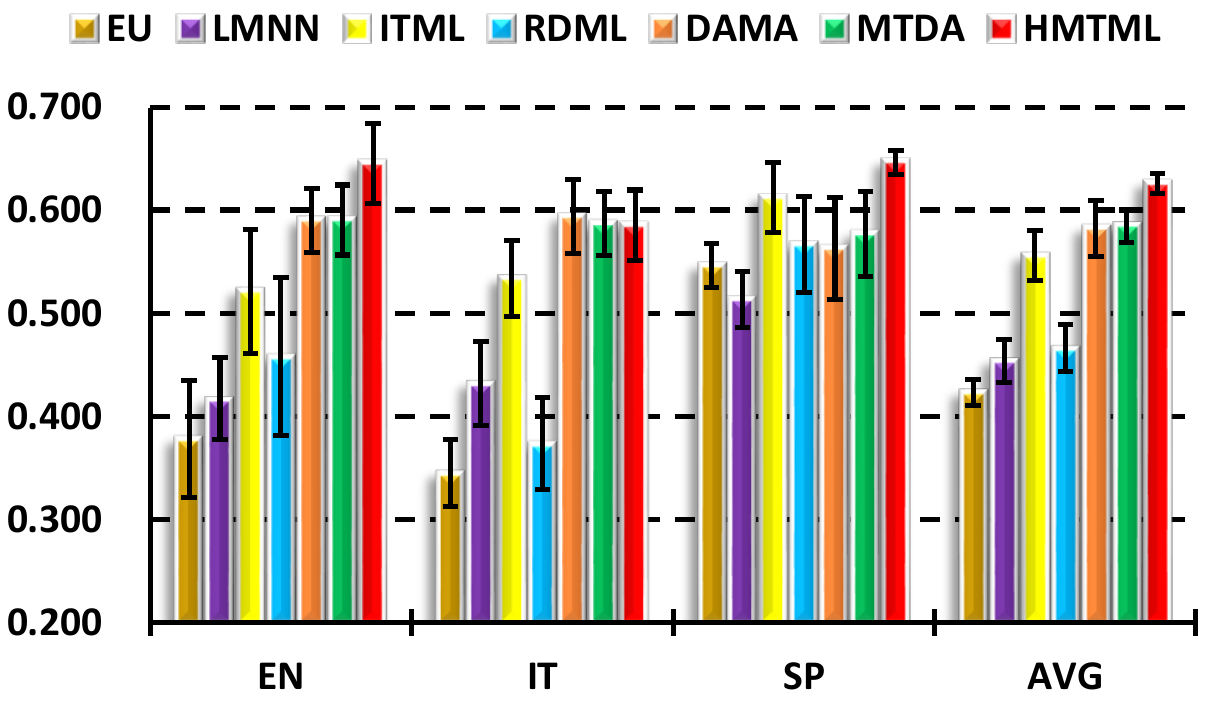}
}
\caption{Individual accuracy and macroF1 score of each domain of the compared methods at their best numbers (of common factors) on the RMLC dataset. ($10$ labeled instances for each category; EN: English, IT: Italian, SP: Spanish, AVG: average.)}
\label{fig:IndvAcc_IndvMacF1_RMLC}
\end{figure}

Fig. \ref{fig:IndvAcc_IndvMacF1_RMLC} compares all different approaches on the individual domains. From the results, we can conclude that: 1) RDML improves the performance in each domain, and the improvements are similar for different domains, since they do not communicate with each other. In contrast, the transfer learning approaches improve much more than RDML in the domains where less discriminative information is contained in the original representations, such as IT (Italian) and EN (English). These results validate the success transfer of the knowledge among different domains, and this is really helpful in metric learning; 2) in the good-performing SP (Spanish) domain, the results of DAMA and MTDA are almost the same as RDML, and sometimes, even a little worse. This phenomenon indicates that in DAMA and MTDA, the discriminative domain can benefit little from the relatively non-discriminative domains. While the proposed HMTML still achieves significant improvements in this case. This demonstrates that the high-order correlation information between all domains is well discovered. Exploring this kind of information is much better than only exploring the correlation information between pairs of domains (as in DAMA and MTDA).

\begin{figure}
\centering
\subfigure{\includegraphics[width=0.48\columnwidth]{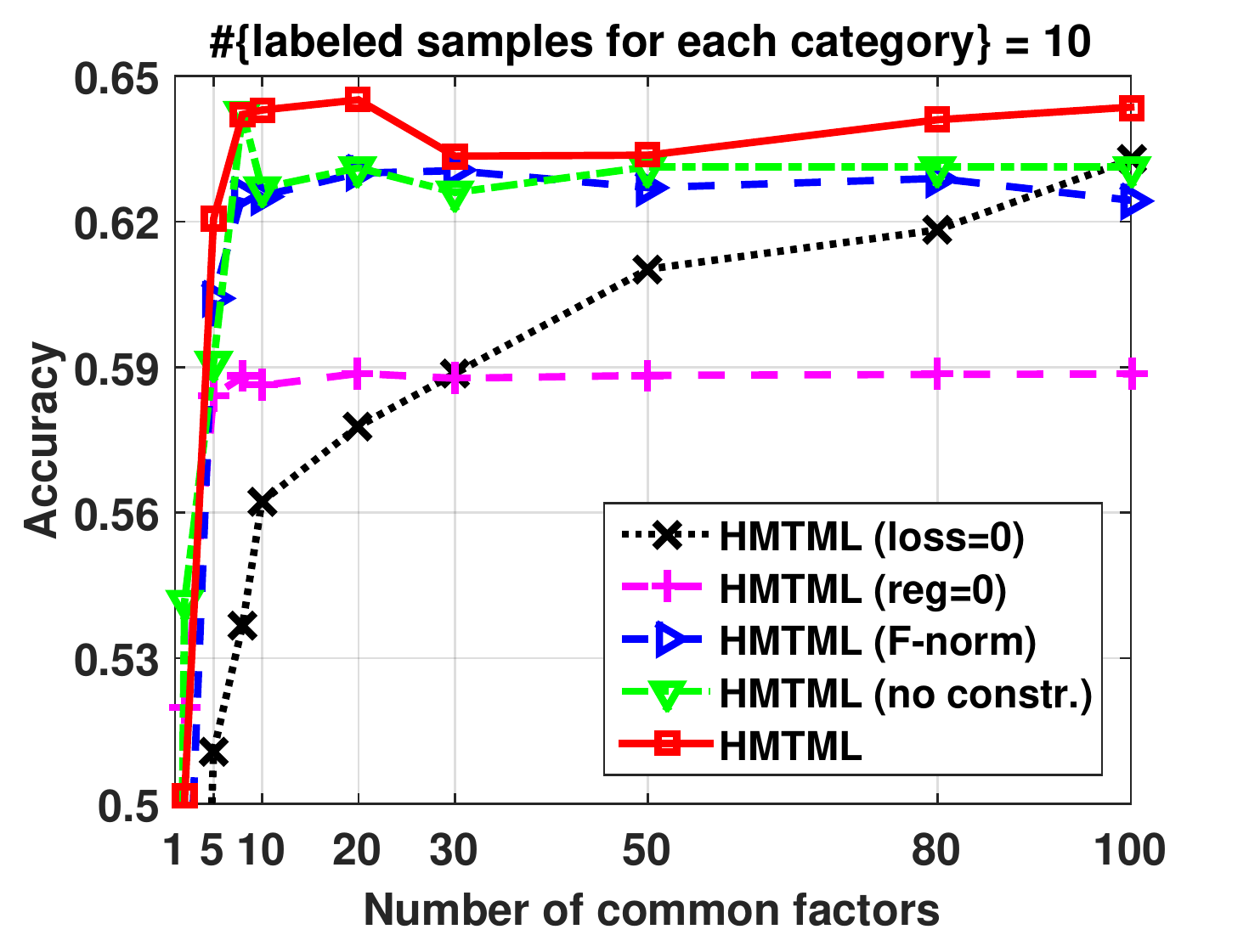}
}
\hfil
\subfigure{\includegraphics[width=0.48\columnwidth]{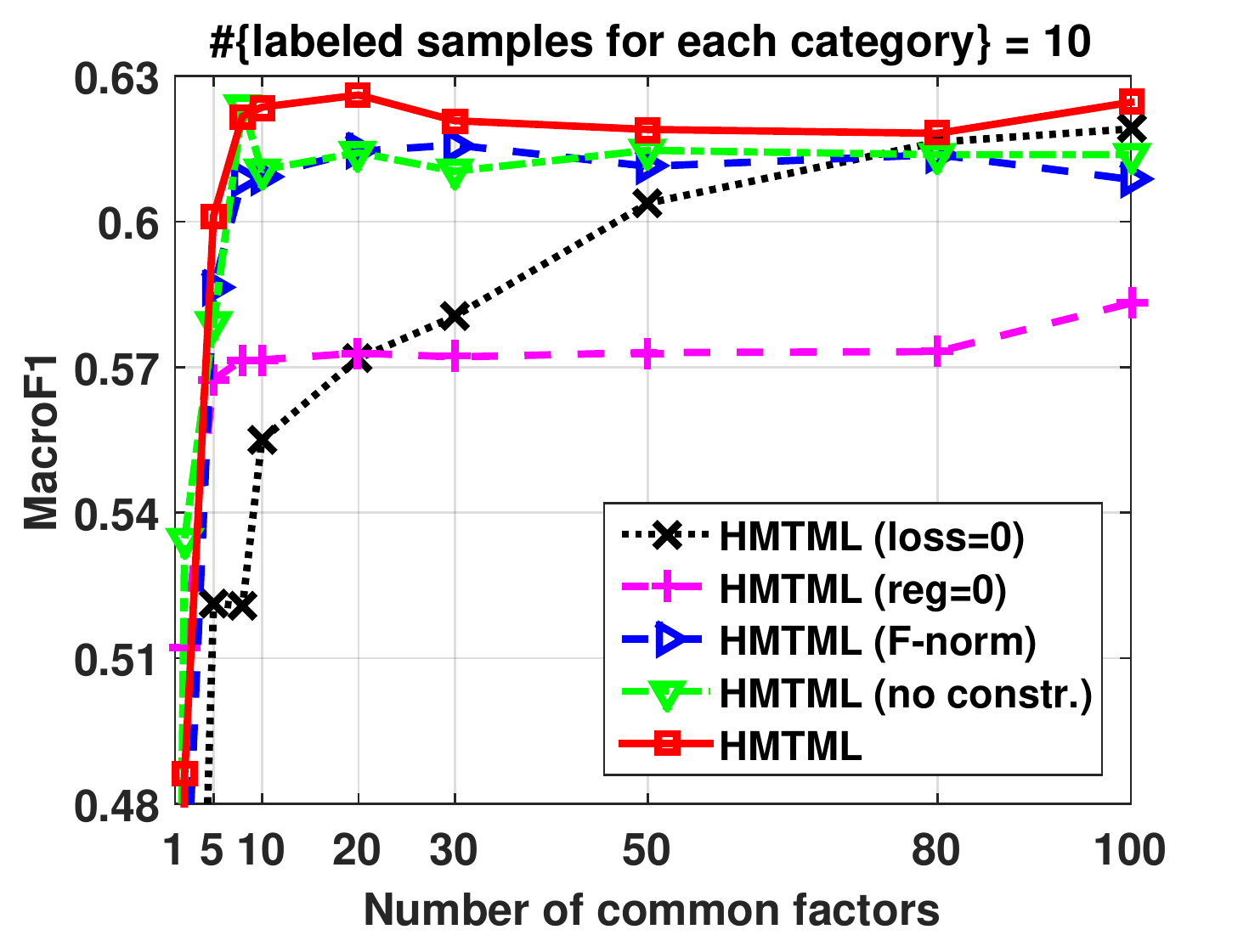}
}
\caption{A self-comparison of the proposed method on the RMLC dataset.}
\label{fig:Acc_MacF1_RMLC_SC}
\end{figure}

\subsubsection{A self-comparison analysis}

To see how the different design choices affect the performance, we conduct a self-comparison of the proposed HMTML. In particular, we compare HMTML with the following sub-models:
\begin{itemize}
  \item \textbf{HMTML (loss=0):} dropping the log-loss term in (\ref{eq:HMTML_Reformulation}), and this amounts to solving just the optimization problem described in (\ref{eq:HMTML_REG}).
  \item \textbf{HMTML (reg=0):} setting the hyper-parameter $\gamma = 0$ in (\ref{eq:HMTML_Reformulation}). Thus the sparsity of the factors is encouraged, whereas the base classifiers are not used.
  \item \textbf{HMTML (F-norm):} replacing the $l_1$-norm with the Frobenius norm in (\ref{eq:HMTML_Reformulation}). The optimization becomes easier since the gradient of the Frobenius norm can be directly obtained without smoothing.
  \item \textbf{HMTML (no constr.):} dropping the non-negative constraints in (\ref{eq:HMTML_Reformulation}). That is, the operator $\pi[x]$ in (\ref{eq:PGM_Update_Rule}) is not performed in the optimization.
\end{itemize}
The results are shown in Fig. \ref{fig:Acc_MacF1_RMLC_SC}, where $10$ labeled instances are selected for each category. From the results, we can see that solving only the proposed tensor-based divergence minimization problem (\ref{eq:HMTML_REG}) can achieve satisfactory performance when the number of common factors is large enough. Both the accuracy and macroF1 score at the best number of common factors are higher than those of HMTML (reg=0), which joint minimizes the empirical losses of all domains without minimizing their divergence in the subspace. HMTML (reg=0) is better when the number of common factors is small and the performance is steadier. Therefore, both the log-loss and divergence minimization terms are indispensable in the proposed model.
HMTML (F-norm) is worse than the proposed HMTML that adopts the $l_1$-norm. This demonstrates the benefits of enforcing sparsity in metric learning, as suggested in the literatures \cite{GJ-Qi-et-al-ICML-2009, W-Liu-et-al-KDD-2010, KZ-Huang-et-al-KIS-2011}.
HMTML (no constr.) is worse than HMTML at most dimensions, although their best performance are comparable. This may be because the non-negative constraints help to narrow the hypothesis space and thus control the model complexity.

\begin{figure*}
\centering
\subfigure[]{\includegraphics[width=0.6\columnwidth]{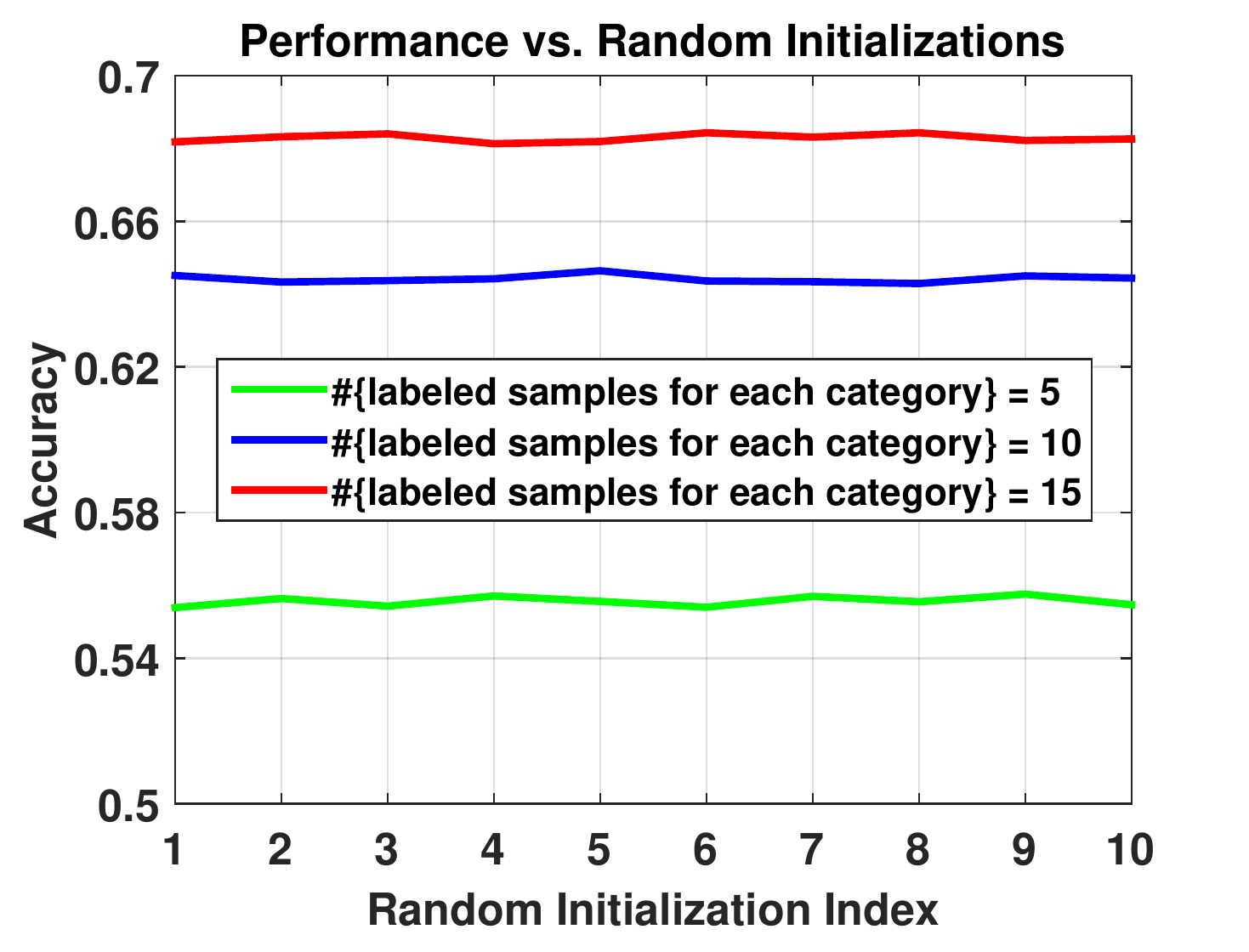}
\label{subfig:RandomInit}
}
\hfil
\subfigure[]{\includegraphics[width=0.6\columnwidth]{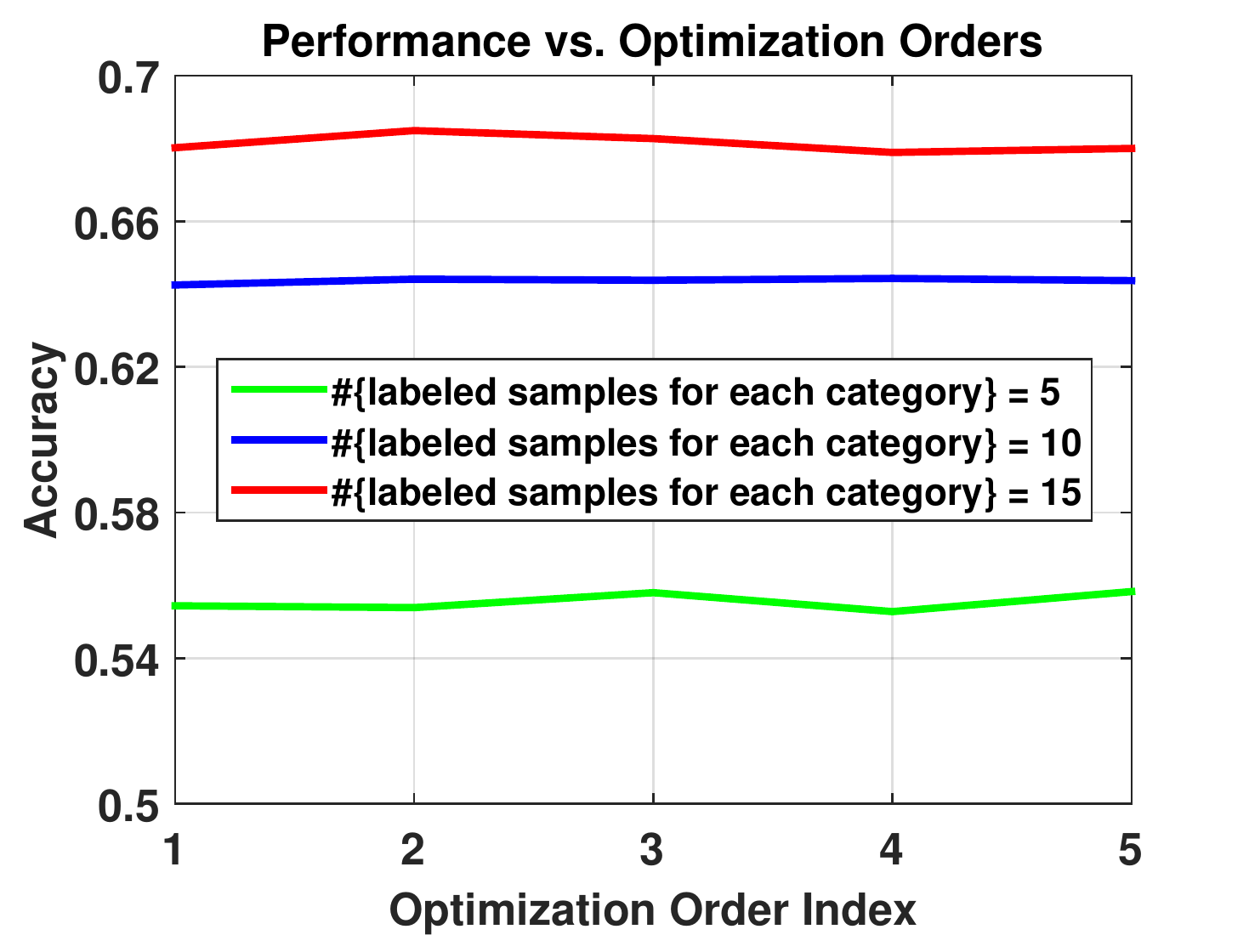}
\label{subfig:OptOrder}
}
\hfil
\subfigure[]{\includegraphics[width=0.6\columnwidth]{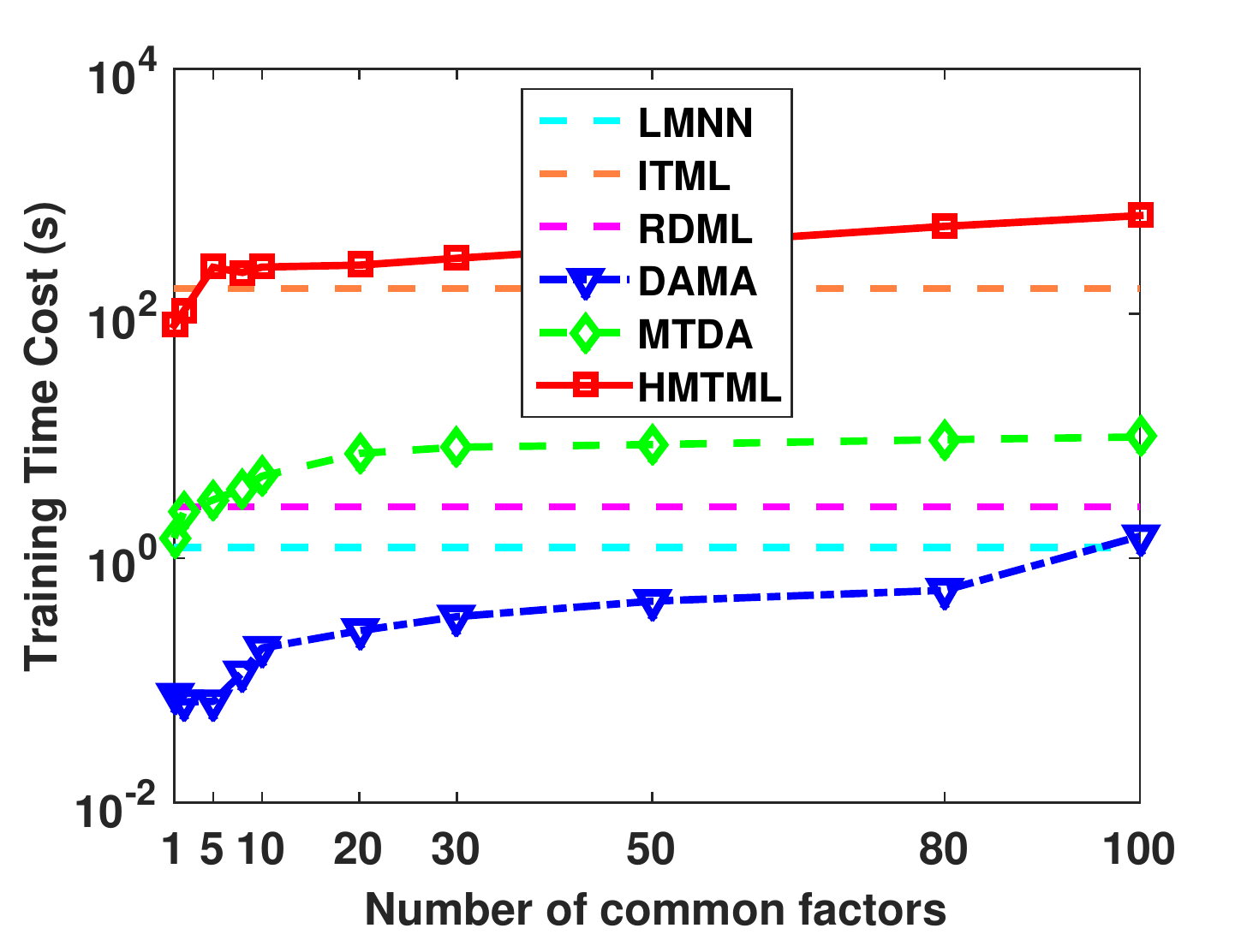}
\label{subfig:TCst_vs_Dim_RMLC}
}
\caption{Empirical analysis: (a) Performance in accuracy vs. random initialized $\{ U_m \}$. The proposed model is insensitive to different initializations; (b) Performance in accuracy vs. order of optimizing $\{ U_m \}$. The proposed model is insensitive to different optimization orders; (c) Training time cost vs. number of common factors on the RMLC dataset ($10$ labeled instances for each category).}
\label{fig:Empirical_Analysis}
\end{figure*}

\subsubsection{Sensitivity analysis w.r.t. different initializations and optimization orders}

We can only obtain a local minimum since the proposed formulation (\ref{eq:HMTML_Reformulation}) is not joint convex with respect to all the parameters. However, the proposed method can achieve satisfactory performance using only random initializations. Fig. \ref{subfig:RandomInit} shows the model is insensitive to different initializations of the parameter set $\{ U_m \}$ and demonstrates the effectiveness of the obtained local optimal solution.

In addition, Fig. \ref{subfig:OptOrder} shows that the proposed model is also insensitive to the different optimization orders of the parameter set $\{ U_m \}$, and demonstrates that the order of optimizing $\{ U_m \}$ would not affect the performance.

%

\subsubsection{Empirical analysis of the computational complexity}

The training time of different approaches can be found in Fig. \ref{subfig:TCst_vs_Dim_RMLC}. All the experiments are performed on the computer with the following specifications: $2.9$ GHz Intel Xeon (8 cores) with Matlab R2014b. The results indicate that the proposed HMTML is comparable to ITML when $r$ (number of common factors) is small, and slower than all other approaches when many common factors are required. The cost of HMTML is not strictly quadratic w.r.t. $r$ because the number of iterations needed for algorithm termination changes (for different $r$). Since the optimal number $r$ is usually not very large, the time cost of the proposed method is tolerable. It should be noted that the test time for different methods are almost the same because the sizes of the obtained metrics are the same.



\section{Conclusion}
\label{sec:Conclusion}

A novel approach for heterogeneous metric learning (HMTML) is proposed in this paper. In HMTML, we calculate the prediction weight covariance tensor of multiple heterogeneous domains. Then the high-order statistics among these different domains are discovered by analyzing the tensor. The proposed HMTML can successfully transfer the knowledge between different domains by finding a common subspace, and make them help each other in metric learning. In this subspace, the high-order correlations between all domains are exploited. An efficient optimization algorithm is developed for finding the solutions. It is demonstrated empirically that exploiting the high-order correlation information achieves better performance than the pairwise relationship exploration in traditional approaches.


It can be concluded from the experimental results on three challenging datasets: 1) when the labeled data is scarce, learning the metric for each individual domain may deteriorate the performance. This issue can be alleviated by jointly learning the metrics for multiple heterogeneous domains. This is in line with the multi-task learning literature; 2) the shared knowledge of different heterogenous domains can be exploited by heterogeneous transfer learning approaches. Each domain can benefit from such knowledge if the discovered common factors are appropriate. We show that it is particularly helpful to utilize the high-order statistics (correlation information) in discovering such factors.

The proposed algorithm is particularly suitable for the case when we want to improve the performance of multiple heterogeneous domains, where the labeled samples are scarce for each of them. Because the learned transformation is linear for each domain, the proposed method is very effective for text analysis based applications (such as document categorization), where the structure of the data distribution is usually linear. For the image data, we suggest adding a kernel PCA (KPCA) preprocessing, which may help improve the performance to some extent.

The relative higher computational cost may be the main drawback of the proposed HMTML when comparing with other heterogeneous transfer learning approaches. In the future, we will develop efficient algorithm for optimization. For example, the technique of parallel computing will be introduced to accelerate the optimization. In addition, to deal with complicated domains, we intend to learn nonlinear metrics by extending the presented HMTML.
\ifCLASSOPTIONcaptionsoff
  \newpage
\fi



\bibliographystyle{IEEEtran}
\bibliography{./TNNLS-2015-P-5907}
\end{document}